\documentclass{article}
\usepackage{dart_preprint,times}

\usepackage{amsmath,amsfonts,bm}

\def\eqref#1{equation~\ref{#1}}

\def\1{\bm{1}}

\DeclareMathAlphabet{\mathsfit}{\encodingdefault}{\sfdefault}{m}{sl}
\SetMathAlphabet{\mathsfit}{bold}{\encodingdefault}{\sfdefault}{bx}{n}

\usepackage{float}
\usepackage{hyperref}
\hypersetup{
  hidelinks,
  pdfauthor={Shihong Li, Juntao Xu, JinCao, Maowen Tang, Jun Huang, Jintao Li},
  pdftitle={DART: Distillation-Aware Reparameterization for Training-Free LoRA Reuse in Few-Step Video Diffusion Models}
}
\usepackage{url}
\usepackage{amsmath,amssymb,amsthm}
\usepackage{graphicx}
\usepackage{booktabs}
\usepackage{multirow}
\usepackage{xcolor}
\usepackage{colortbl}
\usepackage{array}
\usepackage{pifont}
\usepackage{flafter}
\usepackage{placeins}
\usepackage{wrapfig}
\usepackage{needspace}

\theoremstyle{definition}
\newtheorem{definition}{Definition}

\newtheorem{proposition}{Proposition}

\title{\centering\normalfont\large\hyphenpenalty=10000\exhyphenpenalty=10000
DART: Distillation-Aware Reparameterization for Training-Free\\
LoRA Reuse in Few-Step Video Diffusion Models\par}
\author{
\makebox[0.96\textwidth][c]{\normalfont\small
Shihong Li$^{1,*}$ \enspace Juntao Xu$^{2,*}$ \enspace JinCao$^{3}$ \enspace
Maowen Tang$^{1,4}$ \enspace Jun Huang$^{1}$ \enspace Jintao Li$^{1}$}\\[3pt]
\makebox[0.96\textwidth][c]{\normalfont\scriptsize
$^{1}$University of Electronic Science and Technology of China \enspace
$^{2}$Tsinghua University \enspace $^{3}$Harbin Institute of Technology \enspace
$^{4}$Tencent}\\[-1pt]
\makebox[0.96\textwidth][c]{\normalfont\scriptsize $^{*}$Equal contribution.}
}

\newcommand{\rank}{\operatorname{rank}}

\newcommand{\Null}{\operatorname{Null}}

\newcommand{\cmark}{\ding{51}}
\newcommand{\xmark}{\ding{55}}
\newcommand{\firstresult}[1]{\cellcolor{RankFirst}#1}
\newcommand{\secondresult}[1]{\cellcolor{RankSecond}#1}
\newcommand{\thirdresult}[1]{\cellcolor{RankThird}#1}
\newcommand{\tablefont}{\small}
\definecolor{RankFirst}{HTML}{D5E8D4}
\definecolor{RankSecond}{HTML}{FFF2CC}
\definecolor{RankThird}{HTML}{DAE8FC}

\begin{document}

\maketitle
\fancyhead{}

\begin{abstract}
Few-step distillation reduces the inference cost of image-to-video generation, but directly reusing LoRA adapters trained for long denoising trajectories can weaken their intended effects and degrade video quality. We observe that adapters with similar measured static parameter geometry can behave differently under the shortened target schedule, motivating response-aware transfer. We propose DART, a training-free reparameterization method that transports source LoRAs into aligned coordinates through a low-rank distillation bridge. Paired forward evaluations measure channel-level incremental responses under the target schedule to fit coefficients that calibrate response direction, source-relative magnitude, and timestep allocation. Coordinate transport establishes update directions, while calibration adapts their contributions, requiring neither source training videos nor backpropagation. Experiments across multiple distilled I2V models demonstrate improved generation quality and aggregate functional retention over direct reuse. Further analyses show that coordinate transport complements response calibration, with adapter-level benefits encompassing both functional preservation and reduced negative transfer.
\end{abstract}

\section{Introduction}

Few-step image-to-video (I2V) generation offers a practical route to faster video creation, but its usefulness also depends on retaining the customizations learned for existing models. Step distillation and consistency training reduce inference cost by shortening the denoising process~\citep{salimans2022progressive,song2023consistency}, while Low-Rank Adaptation (LoRA) provides lightweight control over appearance and motion~\citep{hu2022lora}. These advances do not automatically compose: a LoRA trained for the original model may lose its intended effect or degrade video quality when attached to a distilled target. Retraining each adapter for every target repeats the cost of customization and may be infeasible when the source training videos are unavailable. Reusing existing LoRAs without retraining is therefore an important deployment problem for few-step I2V models.

A natural approach is to examine whether the adapter remains compatible with the target in weight space. CASA relates transfer failure to spectral-routing interference between distillation and LoRA updates and addresses it through spectral arbitration~\citep{wang2026casa}. This perspective establishes the importance of parameter geometry, but leaves the adapter's behavior during target inference to be understood. Our motivating observation is that adapters with similar measured static geometry can exhibit substantially different functional outcomes after transfer. Static compatibility summaries alone therefore provide an incomplete account of whether the target will retain an adapter's intended effect.

The distinction arises because an adapter acts through a denoiser, not independently of it. Its effect depends on the model weights and on the latent state, timestep, prompt, and first-frame image at which it is applied. Distillation changes the model and the trajectory on which the update operates, so retaining the update's rank or energy does not ensure the same effect during generation. This motivates examining the adapter's \emph{incremental response}: the change in denoiser output induced by the adapter under fixed input conditions. It also explains why good video quality alone is insufficient evidence of successful reuse. Attenuating an incompatible update can reduce generation artifacts while suppressing the customization itself. Transfer must therefore consider how the update acts under target conditions, alongside its static parameter compatibility.

We propose DART, a distillation-aware reparameterization method that couples coordinate transport with target-schedule response calibration. The first design consideration is that calibration can only combine the channel directions it is given. Adjusting coefficients alone leaves those directions fixed, even when distillation has changed the model's coordinates. DART therefore constructs a bridge from a low-rank approximation of the source-to-target weight change and aligns clustered singular subspaces between the source and the bridge. This alignment transports the source LoRA before decomposing it into rank-one channels, providing calibration with directions adapted to the selected distillation change.

Coordinate alignment still does not determine how strongly these channels should contribute at the target denoising timesteps. A single adapter-strength multiplier adjusts all channels together and cannot independently correct their relative contributions. DART instead uses paired forward evaluations on a small probe set to measure channel-level incremental responses under the target schedule. These measurements guide a regularized solve for fixed channel coefficients that calibrate response direction, source-relative magnitude, and allocation across timesteps. Because limited probes cannot distinguish every channel combination, regularization retains the transported initialization in unobserved directions. Transport thus determines which directions are available, while calibration determines how to combine their measured responses. The conversion requires neither source training videos nor denoiser backpropagation, and the resulting adapter can be loaded directly into the distilled target.

Our contributions are threefold.
\begin{itemize}
\item We identify a gap between measured static compatibility and schedule-dependent adapter behavior, motivating incremental responses as a guide for LoRA reuse in distilled I2V models.
\item We introduce DART, a training-free method that combines bridge-based coordinate transport with channel-level, forward-only response calibration, addressing both the available update directions and their contributions under the target schedule.
\item Experiments across multiple distilled I2V models show improved aggregate quality and functional retention over direct reuse. Further analysis supports the complementary roles of transport and calibration, with benefits spanning functional preservation and mitigation of negative transfer.
\end{itemize}

\section{Related Work}

\subsection{Parameter-efficient customization}
Low-Rank Adaptation (LoRA) represents a task update with two low-rank factors while keeping the base model fixed, following a broader line of parameter-efficient adapters and prompt-based updates~\citep{houlsby2019adapters,li2021prefix,lester2021prompt,hu2022lora}. Video customization methods extend this lightweight interface to appearance, subject, and motion: Tune-A-Video and AnimateDiff adapt image diffusion models for video generation, while MotionDirector and VMC learn motion-specific controls~\citep{wu2023tuneavideo,guo2024animatediff,zhao2024motiondirector,jeong2024vmc}. These approaches learn or attach an adapter for a compatible model and denoising trajectory; DART addresses the complementary deployment problem of reusing that adapter after distillation changes the target coordinates and trajectory, without retraining.

\subsection{Few-step video generation}
Progressive distillation, consistency training, and fast ODE solvers shorten diffusion trajectories and reduce inference cost~\citep{salimans2022progressive,song2023consistency,luo2023lcm,lu2022dpmsolver}. Video generation spans autoregressive transformers such as CogVideo~\citep{hong2022cogvideo} and diffusion models such as Imagen Video, Make-A-Video, and ModelScopeT2V~\citep{ho2022imagenvideo,singer2022makeavideo,wang2023modelscope,ho2022video}, with Wan and HunyuanVideo providing representative open video foundation models~\citep{wanteam2025wan,kong2024hunyuanvideo}. VBench evaluates video quality~\citep{huang2024vbench}, while VBench++ adds image-conditioning metrics through VBench-I2V~\citep{huang2024vbenchpp}. These works improve the speed or measurement of generation, whereas DART studies whether a customization learned for a long trajectory remains functional on a shortened target schedule.

\subsection{Data-free adapter transfer}
Data-free adapter transfer avoids source training data. CASA provides a spectral-arbitration baseline for video LoRA reuse by relating failure to routing interference between distillation and LoRA updates~\citep{wang2026casa}. Its weight-space analysis motivates examining parameter compatibility. Our observations raise an additional question: how does the update affect denoiser outputs at target states and timesteps? DART combines coordinate adaptation with target-schedule response calibration and evaluates quality separately from signed functional effects. We compare against CASA in its original training-free formulation. DART is an alternative transfer framework, rather than a modification of CASA; the comparison does not isolate the relative merits of the two coordinate-transport procedures.

Appendix~\ref{app:extended_related_work} distinguishes weight-space transfer and merging, adapter reparameterization and mapping, and function-space response matching. It positions DART as target-schedule-conditioned, channel-level, forward-only calibration for distilled image-to-video (I2V) LoRA reuse.

\section{Exploratory Study of LoRA Reuse}
\label{sec:exploration}

CASA studies transfer in weight space~\citep{wang2026casa}: full-model and LoRA updates largely preserve singular values, while structured rotations within singular subspaces create routing interference. This establishes static spectral compatibility as an important part of data-free transfer.

We ask how measured static compatibility relates to functional behavior after distillation. Four-anchor evaluation separates functional effects from quality; static geometry and response summaries provide descriptive context. A global-scale sweep supplies an additional baseline. We use a 40-step Wan2.2-I2V-A14B source and its four-step target. For each adapter, eight prompts are crossed with eleven first-frame reference images, yielding 88 videos per method. Of the eight initially evaluated LoRAs, six are retained; LoRAs 02 and 08 are excluded because source gains are near zero. Appendix~\ref{app:lora_inventory} reports the complete inventory and functional roles. Figure~\ref{fig:geometry_response} contrasts static geometry with target-schedule responses; Appendix~\ref{app:exploration_fig} provides the anchor summary and global-scale sweep.

\subsection{Four anchors separate function from quality}
\label{sec:four_anchor_main}

For adapter $i$, let $A_i$ be Source without LoRA, $B_i$ Source with the original LoRA, $C_i$ Target without LoRA, and $D_{i,m}$ Target with transfer method $m$. For a functional score $s$, the adapter-level recovery is
\begin{equation}
\mathrm{FGR}_{i,m}
=\frac{\bar{s}(D_{i,m})-\bar{s}(C_i)}
{\bar{s}(B_i)-\bar{s}(A_i)}.
\label{eq:fgr_main}
\end{equation}
The bar pools the eight prompts before the ratio. The proxy is selected from source anchors before target outcomes: CSD~\citep{somepalli2024style} for LoRA 01 and VideoMAE~\citep{tong2022videomae} for LoRA 03--07. Target no-LoRA quality averages $Q_C=0.9236$, versus $Q_D=0.9029$ for Direct, while coordinate defects remain in the narrow $2.635$--$2.652$ range (units of $10^{-3}$). Functional and quality outcomes are therefore not interchangeable.

\subsection{Static geometry and target-schedule responses}

\begin{wrapfigure}{R}{0.64\textwidth}
\centering
\includegraphics[width=\linewidth]{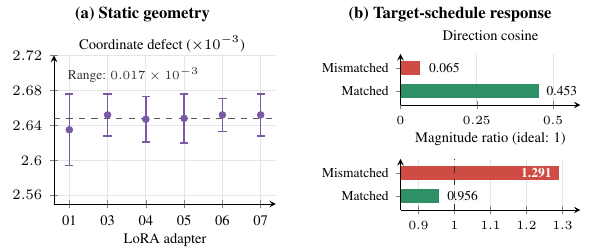}
\caption{\textbf{Static geometry and target-schedule responses.} (a) Coordinate defects vary little across adapters (mean $\pm$ layer SD; ten layers each). (b) Branch-mismatched responses have lower direction cosine and larger magnitude ratios than matched responses, before DART.}
\label{fig:geometry_response}
\end{wrapfigure}

After cluster-aware alignment, coordinate defect is about $0.27\%$ across ten layers per adapter, with subspace alignment $0.999553$ and spectral alignment $0.999999658$. Figure~\ref{fig:geometry_response}(a) shows the narrow range of coordinate defects across the six retained adapters. Despite this similarity, direct $Q_{\rm joint}$ spans $0.8045$--$0.9329$ and direct FGR spans $-1.2714$--$0.3397$ (Appendix Table~\ref{tab:exploration_anchor_geometry}). These measured geometry summaries do not distinguish the observed functional outcomes in this adapter set.

Figure~\ref{fig:geometry_response}(b) examines source-to-target responses before DART. The branch-mismatched observations have lower response-direction cosine and larger target-to-source magnitude ratios than the matched observations (Table~\ref{tab:response_diagnostics}). This descriptive difference motivates measuring responses under target conditions, alongside static coordinate compatibility. It does not establish a causal effect of schedule shortening or guarantee functional recovery. Together, the two panels motivate considering both coordinate transport and target-schedule response calibration.

\subsection{Global scaling as a supporting comparison}

A global scale is a simple data-free baseline. All scales use the same evaluation cases, seeds, and scoring protocol. The endpoints are target no-LoRA at $\alpha=0$ ($Q_{\rm joint}=0.9236$, $R_{\rm LoRA}=0$) and Direct at $\alpha=1$ ($0.9029$, $-0.4644$). Reducing the scale to $0.75$ raises retention to $0.1178$ but lowers quality to $0.9011$. The no-LoRA endpoint has the highest quality but no functional recovery, while $0.75$ has the highest retention among the tested scales. These sampled results expose a quality--function trade-off, rather than establish that every possible global scale is inadequate. Appendix Figure~\ref{fig:scale_sweep} visualizes this trade-off, and Table~\ref{tab:app_scale} reports the full sweep.

\section{DART}
\label{sec:method}

\subsection{Problem and method overview}

We transfer a source LoRA update $C$ from source weights $W_s$ to distilled target weights $W_t$. The notation describes one adapted layer, with the same construction applied to the selected layers. The target uses $m$ denoising times $\tau_1,\ldots,\tau_m$. A probe $x$ contains a latent state, a prompt, and a reference image. Because distillation changes both the weights and the trajectory, we consider coordinate compatibility and measured responses under these target conditions.

\begin{figure*}[t]
\centering
\includegraphics[width=\textwidth]{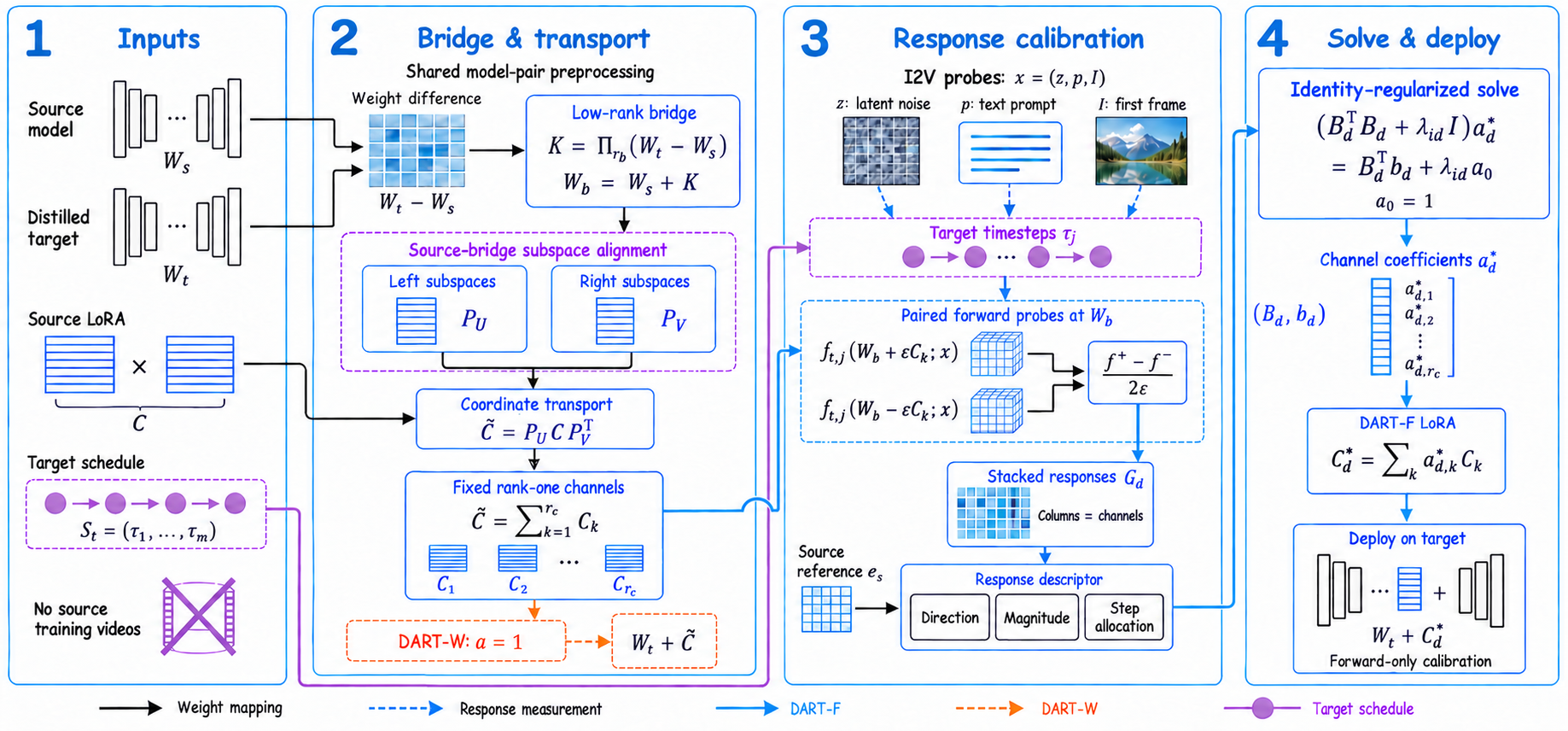}
\caption{\textbf{DART procedure.} Coordinate transport, target-schedule response calibration, and deployment on distilled target weights.}
\label{fig:method}
\end{figure*}

Figure~\ref{fig:method} outlines DART-F. A low-rank bridge supports coordinate transport and paired channel-response probing under the target schedule. The stacked response matrix $G_d$ yields descriptors for an identity-regularized solve; the calibrated update is deployed on $W_t$. Appendix~\ref{app:theory} defines the response space and details the matrix construction, assumptions, and proofs.

\subsection{Bridge construction and coordinate transport}

The bridge isolates a selected part of the distillation change for coordinate alignment. Let $\Pi_{r_b}$ denote the selected rank-$r_b$ approximation, where $r_b$ is the bridge rank. We define the bridge weights $W_b$ and the residual drift $R$ by
\begin{equation}
W_b=W_s+\Pi_{r_b}(W_t-W_s),
\qquad
R=W_t-W_b.
\label{eq:bridge_main}
\end{equation}
The bridge is used for alignment and probing; deployment uses $W_t$, so residual drift $R$ remains relevant.

Clustered singular subspace alignment gives a left map $P_U$ and a right map $P_V$. They align source and bridge directions while allowing rotations within clusters of nearly equal singular values. The transported adapter is
\begin{equation}
\widetilde C=P_UCP_V^\top.
\label{eq:transport_main}
\end{equation}
Appendix~\ref{app:transport_construction} defines the maps and their support conditions. These conditions distinguish a coordinate change from a projection that can discard part of the adapter. For a fixed model pair and alignment configuration, bridge construction, model-weight SVD, clustering, and coordinate-map construction are cached once; map application and response calibration remain adapter-specific.

\begin{proposition}[Transport invariants]
\label{prop:invariants}
If $P_U$ and $P_V$ are isometries on the column and row spaces occupied by $C$, then $\widetilde C$ and $C$ have the same rank, nonzero singular values, and Frobenius norm.
\end{proposition}

Under the stated support conditions, transport preserves rank and energy. These invariants do not imply preservation of the generated adapter effect.

\subsection{Channels and measured responses}

A single multiplier can change only the strength of the complete update. To allow distinct corrections, we decompose $\widetilde C$ into $r_c$ fixed rank-one matrices $C_k$ and assign one coefficient $a_k$ to each channel. Here $a$ is the coefficient vector and $a_0$ is the vector of ones. The calibrated update is
\begin{equation}
C_t(a)=\sum_{k=1}^{r_c}a_kC_k,
\qquad
C_t(a_0)=\widetilde C,
\qquad
a_0=\mathbf{1}.
\label{eq:channels_main}
\end{equation}
The channel count $r_c$ defines the coefficient space, while bridge rank $r_b$ selects the distillation change used for alignment.

We fit coefficients to their measured responses. Let $f_{t,j}(W;x)$ be the denoiser output at target time $\tau_j$ and probe $x$; for update $A$, define its effect relative to the bridge baseline
\begin{equation}
\mathcal R_{t,j}(A;x)
=f_{t,j}(W_b+A;x)-f_{t,j}(W_b;x).
\label{eq:response_main}
\end{equation}
Paired evaluations at $W_b\pm\epsilon C_k$ estimate each local channel response. Stacking probes and target times gives the measured operator $\widehat{\mathcal G}_d$ for target configuration $d$; Appendix~\ref{app:probe_operator} gives the finite-difference construction.

\begin{definition}[Schedule-visible equivalence]
Two coefficient vectors are equivalent on the measured target configuration when $\widehat{\mathcal G}_d a=\widehat{\mathcal G}_d a'$. The rank of $\widehat{\mathcal G}_d$ is the schedule-visible rank $r_{\rm vis}$.
\end{definition}

Calibration can identify only channel combinations exposed by the selected probes.

\subsection{Response calibration}

We fit direction, source-relative magnitude, and step-allocation descriptors. After weighting and stacking them, $B_d$ maps channel coefficients to descriptors and $b_d$ contains their targets; Appendix~\ref{app:calibration_construction} specifies both.

A positive identity penalty $\lambda_{\rm id}$ discourages unsupported changes from $a_0$. We therefore minimize
\begin{equation}
\mathcal L_d(a)=\|B_da-b_d\|_2^2
+\lambda_{\rm id}\|a-a_0\|_2^2,
\qquad
a_d^\star=\arg\min_a\mathcal L_d(a).
\label{eq:objective_main}
\end{equation}
This quadratic objective has the unique solution
\begin{equation}
a_d^\star=(B_d^\top B_d+\lambda_{\rm id}I)^{-1}
(B_d^\top b_d+\lambda_{\rm id}a_0).
\label{eq:closed_form_main}
\end{equation}
The deployed update is $C_d^\star=C_t(a_d^\star)$ on $W_t$; the positive-definite system requires no denoiser backward pass.

\begin{proposition}[Visible correction and scalar limitation]
\label{prop:calibration_main}
With $\lambda_{\rm id}>0$, the optimal change $a_d^\star-a_0$ has zero projection onto the null space of $\widehat{\mathcal G}_d$. If the desired descriptor $b_d$ lies outside the span of $B_da_0$, every single global scale has strictly positive descriptor error.
\end{proposition}

Unobservable combinations stay at the transported identity. Channel calibration can fit multiple measured descriptor directions, but cannot guarantee attainable targets or exact prediction of generated videos.

Appendix~\ref{app:theory} analyzes this local response model, including regularization bias, residual drift, finite-difference error, and target dependence. These conditional results explain the construction and its limitations; they do not establish semantic recovery or characterize all deployment trajectories.

\paragraph{Variants.}
DART-W uses the transported update with $a=a_0$. DART-C retains the direct channel coordinates and solves Equation~\ref{eq:closed_form_main}. DART-F performs both transport and calibration.

\section{Experiments}
\label{sec:experiments}

\subsection{Experimental setup and metrics}
\label{sec:i2v_setup}

We evaluate image-to-video (I2V) LoRA reuse from a 40-step Wan2.2-I2V-A14B source to its four-step distilled target. Each of eight adapters uses $8$ prompts $\times$ $11$ first-frame images, giving $88$ videos per method. LoRAs 02 and 08 are excluded before target outcomes are inspected because their source-side functional gains are near zero, leaving six adapters: $528$ videos per method and $2{,}640$ across the five methods, excluding anchors. All 88 cases contribute to quality and FGR. Videos use $640\times640$ resolution, 129 frames, 16 fps, and \texttt{bfloat16} inference with LightX2V. Appendices~\ref{app:lora_inventory} and~\ref{app:i2v_reference_images} document adapter roles and reference images.

\paragraph{Comparisons.}
Direct attachment tests unmodified reuse, while CASA is the external training-free transfer baseline. DART-W and DART-C isolate coordinate transport and response calibration; DART-F combines them. The target no-LoRA anchor separates generation quality from adapter-induced effects. Conversion uses no source training videos, and its four calibration probes are distinct from the evaluation cases.

\paragraph{Metrics.}
Following VBench and VBench-I2V~\citep{huang2024vbench,huang2024vbenchpp}, VBench-Q averages six normalized quality dimensions, excluding dynamic degree; I2V-Avg averages two image-conditioning consistency dimensions. Using unrounded paired scores, we give the two groups equal weight:
\begin{equation}
Q_{\rm joint}(m)=\frac{1}{2}\bigl(\mathrm{VBench\mbox{-}Q}(m)+\mathrm{I2V\mbox{-}Avg}(m)\bigr).
\label{eq:qjoint}
\end{equation}
Functional retention uses the source-selected proxy and four anchors from Section~\ref{sec:four_anchor_main}. We pool the 88 cases within each adapter before averaging its signed recovery ratio over $N=6$ adapters:
\begin{equation}
R_{\rm LoRA}(m)=\frac{1}{N}\sum_{i=1}^{N}
\frac{\bar{s}(D_{i,m})-\bar{s}(C_i)}
{\bar{s}(B_i)-\bar{s}(A_i)}.
\label{eq:r_lora}
\end{equation}
A value of one denotes mean source-normalized gain of one, zero denotes zero net normalized gain, and a negative value denotes a negative mean signed effect relative to target no-LoRA. This proxy-based average does not imply recovery for every adapter. Tables highlight the top three values in green, yellow, and blue.

\paragraph{Choice of four probes.}
Controlled high-noise diagnostics on three adapters show that four probes improve held-out nonlinear response agreement with the eight-probe fit over one or two probes, and reduce subset sensitivity relative to two. Six improve stability further. Four therefore provide an empirical computation--stability trade-off, halving probe-dependent forward evaluations relative to eight. This evidence concerns synthetic probes, not robustness to real prompts, reference images, or final-video metrics (Appendix~\ref{app:probe_robustness}).

\subsection{Main results}

DART-F improves both aggregate quality and functional retention over Direct and CASA (Table~\ref{tab:main}). Relative to Direct, $Q_{\rm joint}$ increases from $0.9029$ to $0.9227$, while $R_{\rm LoRA}$ changes from $-0.4644$ to $+0.1349$. Quality approaches the target no-LoRA mean of $0.9236$, while the positive mean signed effect distinguishes the result from simply removing the adapter. DART-F ranks first, including ties, on seven reported dimensions and second on imaging (Appendix Table~\ref{tab:dimensions}). These are aggregate gains: four adapters improve in quality and two decline, with the largest gain on LoRA 01. Section~\ref{sec:adapter_behavior} examines this uneven distribution.

\begin{table*}[!htbp]
\centering
\tablefont
\caption{\textbf{Unified LoRA reuse results.} Aggregate quality and functional retention for direct attachment, CASA, and DART variants.}
\label{tab:main}
\resizebox{\textwidth}{!}{%
\begin{tabular*}{\textwidth}{@{\extracolsep{\fill}}l l cccc@{}}
\toprule
\textbf{Category} & \textbf{Method} & $Q_{\rm joint}\uparrow$ & $R_{\rm LoRA}\uparrow$ & \textbf{VBench-Q}$\uparrow$ & \textbf{I2V-Avg}$\uparrow$ \\
\midrule
\multirow{2}{*}{Baselines} & Direct attachment & \thirdresult{0.9029} & \thirdresult{$-0.4644$} & \thirdresult{0.8502} & \thirdresult{0.9556} \\
& CASA & 0.8918 & $-0.6071$ & 0.8467 & 0.9369 \\
\midrule
\multirow{3}{*}{DART variants} & DART-W & 0.8950 & $-0.5347$ & 0.8465 & 0.9436 \\
& DART-C & \secondresult{0.9200} & \secondresult{$-0.2973$} & \secondresult{0.8697} & \secondresult{0.9702} \\
& \textbf{DART-F} & \firstresult{0.9227} & \firstresult{+0.1349} & \firstresult{0.8732} & \firstresult{0.9721} \\
\bottomrule
\end{tabular*}%
}
\end{table*}

\begin{figure}[H]
\centering
\includegraphics[width=0.72\linewidth]{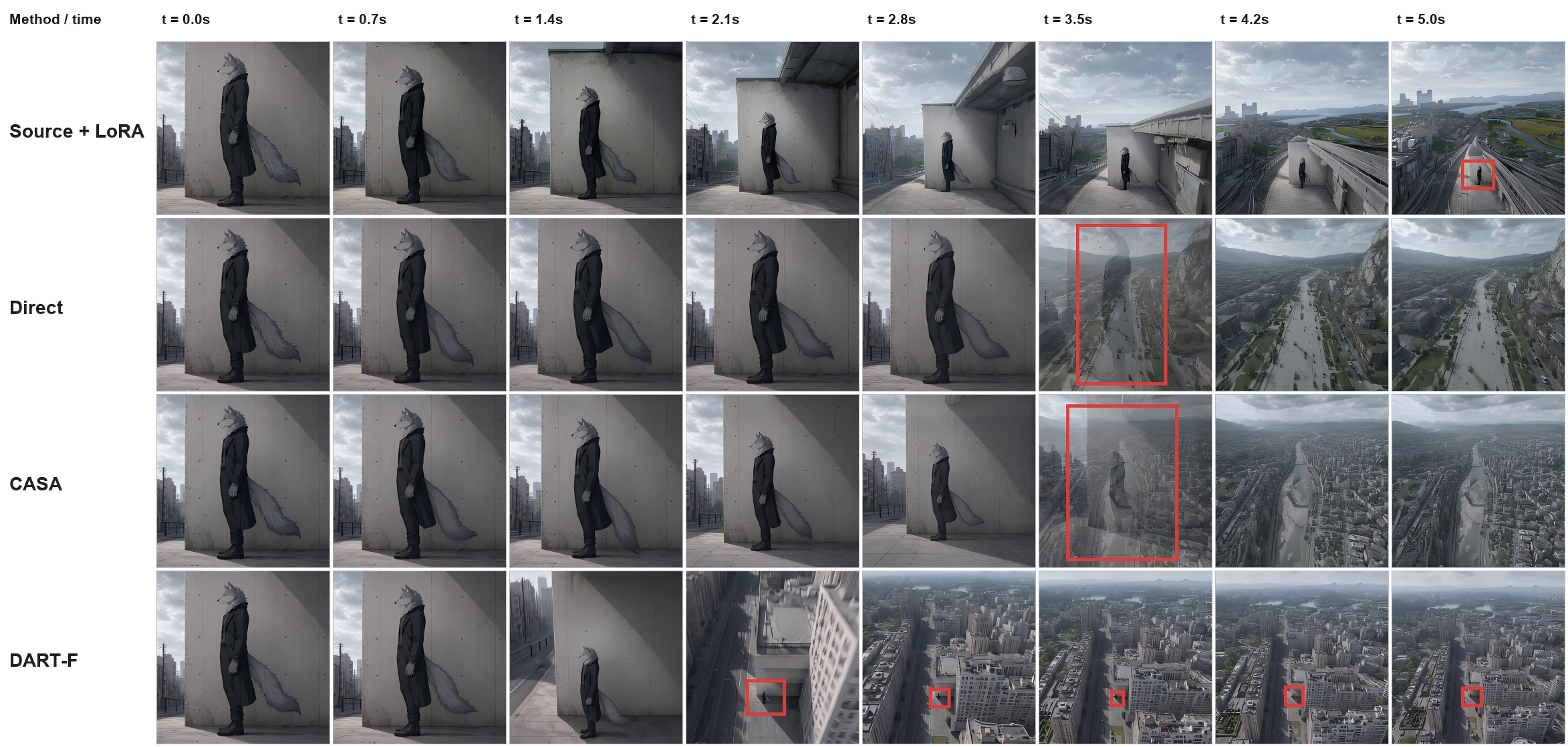}
\caption{\textbf{Representative qualitative comparison.}}
\label{fig:qualitative}
\end{figure}

Figure~\ref{fig:qualitative} presents a representative comparison using an adapter for aerial camera motion. Its prompt is: \emph{The drone shot starts with the main subject in the first frame and gradually ascends upwards, with a smooth and natural transition.} Source with the original LoRA and DART-F show the requested upward movement, whereas Direct and CASA abruptly switch to a distant view with a subject-shaped remnant. This example illustrates visible camera behavior; the adapter-level outcomes in Section~\ref{sec:adapter_behavior} distinguish such observations from aggregate functional recovery.

\subsection{Component analysis}

The component comparison separates quality improvement from functional recovery. Calibration supplies most of the quality gain: DART-C reaches $Q_{\rm joint}=0.9200$, but its mean retention remains negative at $-0.2973$. Combining transport with calibration raises quality to $0.9227$ and changes retention to $+0.1349$ (Table~\ref{tab:main}). Thus, transport's contribution within the full method is clearer on the functional axis than from the additional $0.0027$ quality gain alone. DART-W by itself lowers aggregate quality relative to Direct, so the results support complementary, rather than equal or independently beneficial, contributions (Appendix Table~\ref{tab:ablation}).

LoRA 04 makes this distinction concrete. Calibration alone leaves primary FGR near Direct ($-0.8104$ versus $-0.8459$); transport alone restores a positive effect ($+0.4910$), and the combination reaches $+0.5377$. In the ArcShot example, Direct and DART-C retain the side view, whereas DART-W partially recovers the arc and DART-F most closely approaches the source low-angle view (Appendix Figure~\ref{fig:lora04_arcshot_case}). This case supports adapting the channel directions before calibrating their contributions, without implying that transport benefits every adapter.

\Needspace{17\baselineskip}
\subsection{Additional target evaluation}

\begin{wraptable}{r}{0.60\textwidth}
\centering
\tablefont
\caption{\textbf{Additional target evaluation.} Results on HunyuanVideo 1.5 and CausalWan2.2-I2V-A14B.}
\label{tab:cross}
\setlength{\tabcolsep}{3pt}
\begin{tabular*}{\linewidth}{@{\extracolsep{\fill}}l l cc@{}}
\toprule
\textbf{Target model} & \textbf{Method} & $Q_{\rm joint}\uparrow$ & $R_{\rm LoRA}\uparrow$ \\
\midrule
\multirow{3}{*}{\shortstack[l]{HunyuanVideo\\1.5}}
& Direct & \thirdresult{0.8778} & \thirdresult{$-0.4372$} \\
& CASA & \secondresult{0.8897} & \secondresult{$-0.2447$} \\
& \textbf{DART} & \firstresult{0.9014} & \firstresult{0.0771} \\
\midrule
\multirow{3}{*}{\shortstack[l]{CausalWan2.2-\\I2V-A14B}}
& Direct & \thirdresult{0.8976} & \thirdresult{$-0.4186$} \\
& CASA & \secondresult{0.9026} & \secondresult{$-0.1843$} \\
& \textbf{DART} & \firstresult{0.9117} & \firstresult{0.1286} \\
\bottomrule
\end{tabular*}
\end{wraptable}

We examine whether the primary trend also occurs on two additional distilled targets. On HunyuanVideo 1.5~\citep{wu2025hunyuanvideo15} and CausalWan2.2-I2V-A14B, DART has the highest joint quality and the only positive mean retention among Direct, CASA, and DART (Table~\ref{tab:cross}). Both baselines retain negative mean functional effects on each target, while DART improves the quality--function combination. The agreement with the primary comparison supports reuse across these evaluated model pairs. Conversion is performed for each target separately; these results do not test transfer of one converted adapter to unseen models or schedules.

\subsection{Adapter-level behavior}
\label{sec:adapter_behavior}

\begin{wraptable}{r}{0.58\textwidth}
\centering
\tablefont
\caption{\textbf{Target-schedule response diagnostic.} Pre-DART fixed-state source and target response summaries; count denotes observations.}
\label{tab:response_diagnostics}
\setlength{\tabcolsep}{3pt}
\begin{tabular*}{\linewidth}{@{\extracolsep{\fill}}l c cc@{}}
\toprule
\textbf{Subset} & \textbf{Count} & \shortstack{\textbf{Direction}\\\textbf{cosine}} & \shortstack{\textbf{Magnitude}\\\textbf{ratio}} \\
\midrule
Branch mismatched & 3 & 0.0650 & 1.2912 \\
Branch matched & 9 & 0.4529 & 0.9556 \\
\bottomrule
\end{tabular*}
\end{wraptable}

The response summary in Table~\ref{tab:response_diagnostics} describes the pre-DART mismatch motivating Section~\ref{sec:exploration}, not a post-calibration recovery test; Appendix Table~\ref{tab:app_response} provides the full statistics. To interpret DART's outcomes, we instead examine adapter-level scores alongside the fitted channel coefficients.

LoRAs 04, 05, and 07 have positive primary FGR ($+0.5377$, $+0.5229$, and $+0.2780$) and retain both active and attenuated channels. LoRAs 01, 03, and 06 have all 32 coefficients near zero; their FGR improves over Direct but remains negative. This pattern is consistent with selective functional preservation in the former group and attenuation of incompatible updates in the latter, rather than uniform recovery (Appendix Tables~\ref{tab:app_fgr_detail} and~\ref{tab:app_coeff}).

Quality gains also depend strongly on the adapter. LoRA 01 improves by $0.1213$ in $Q_{\rm joint}$, but excluding it changes the mean gain from $+0.0198$ to $-0.0006$ (Appendix~\ref{app:adapter_heterogeneity}). Its near-zero coefficients and still-negative FGR distinguish mitigation of severe quality degradation from restoration of its intended function. Accordingly, aggregate quality and positive macro retention should be read alongside these per-adapter outcomes.

\subsection{Qualitative results and conversion cost}

\begin{figure}[!h]
\centering
\includegraphics[width=0.9\linewidth]{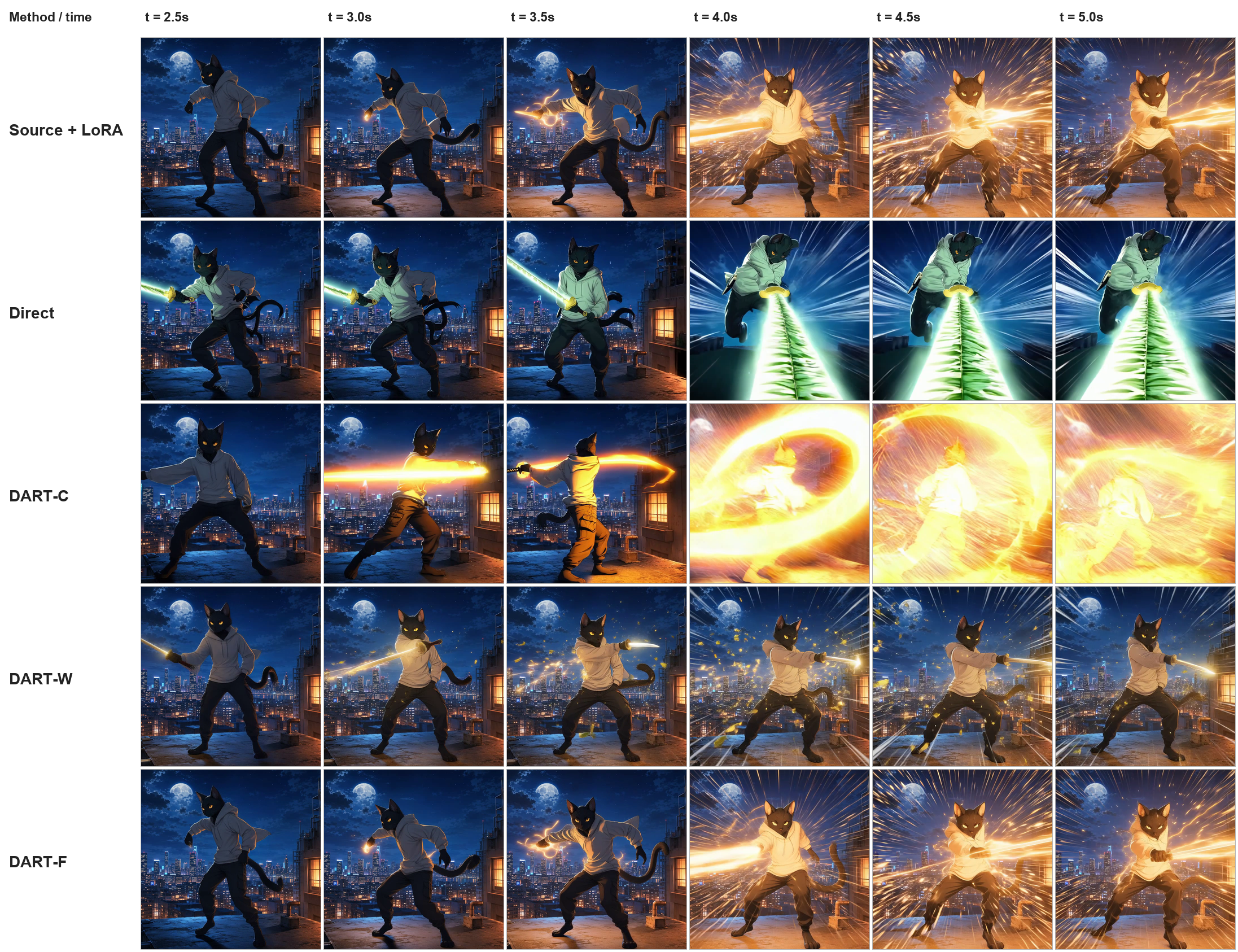}
\caption{\textbf{LoRA 05 (Anime Lore) qualitative component comparison.}}
\label{fig:ablation_case}
\end{figure}

LoRA 05 (Anime Lore) targets cel-animation appearance, anime lighting, and stylized action. Its prompt is: \emph{The subject in the first frame draws a sword in a single sweeping motion, glowing energy trailing the blade, wind kicking up leaves while dramatic anime speed lines converge behind.} Figure~\ref{fig:ablation_case} shows a coherent sword sweep and trailing energy with Source and DART-F. Direct produces an oversized green blade, DART-C amplifies the energy into a bright ring, and DART-W localizes the blade effect. Together with the ArcShot case, this illustrates how the component differences appear in generated frames.

\paragraph{Conversion cost.}
DART-F averages $226.5$ minutes per measured conversion, including $96$ minutes of model-pair preprocessing reusable across LoRAs. Subtracting this shared setup gives an estimated $130.5$ minutes per additional adapter, not a separately timed measurement. Conversion requires no source training videos or backpropagation, but does incur offline computation; these costs exclude target video generation (Appendix Table~\ref{tab:app_efficiency}).

\FloatBarrier

\section{Conclusion}

We study LoRA reuse on few-step distilled video diffusion models and show that static parameter compatibility alone does not fully characterize adapter behavior under a shortened denoising schedule. DART addresses this gap through training-free low-rank coordinate transport and target-schedule response calibration, requiring only forward evaluations and no source training videos. Across the primary Wan2.2 setting and two additional targets, DART improves aggregate quality and signed functional retention over direct reuse in our evaluations. Our results highlight target-schedule functional response as a complementary signal to parameter compatibility for reusing existing LoRAs on accelerated video generation models. Limitations and future directions are discussed in Appendix~\ref{app:limitations}.

\bibliography{dart_references_expanded_20260909,dart_references_related_work_20260917}
\bibliographystyle{dart_references}

\section*{Statement on Language Model Assistance}

The authors used language-model assistance for drafting, editing and table formatting. The authors are responsible for the research ideas, mathematical claims, implementation, experiments, citations, and final verification.

\section*{Reproducibility Statement}

Section~\ref{sec:i2v_setup} specifies the source and target models, generation resolution, frame count, inference precision, evaluation metrics, and the number of prompt--image cases used for every adapter and method. Appendix~\ref{app:lora_inventory} documents the source LoRA inventory and functional roles; Appendix~\ref{app:theory} gives the response operator, transport construction, calibration objective, assumptions, and proofs; and Appendix~\ref{app:probe_robustness} reports probe construction, random seeds, numerical precision, calibration weights, subset enumeration, and finite-difference checks for the robustness study. Appendices~\ref{app:details} and~\ref{app:additional} provide the complete adapter-level, anchor, ablation, quality-dimension, functional-retention, coefficient, and conversion-cost results. All 88 prompt--reference-image cases per adapter and method contribute to the reported primary quality and FGR statistics, with common cases, seeds, and scoring protocols used for method comparisons.

\clearpage
\appendix
\raggedbottom

\section{Extended Related Work and Positioning}
\label{app:extended_related_work}

DART studies reuse of an existing LoRA on a distilled image-to-video (I2V) model, conditioned on a reference image and a text prompt. The related literature addresses parameter compatibility, adapter representation, and functional alignment. These are overlapping perspectives: a mapping can support merging, and a weight update can be selected using response measurements. The distinction below concerns what each method transforms or matches, rather than assigning every method to a mutually exclusive category. Results from text-to-image generation or language modeling supply methodological context, not direct evidence of I2V transfer performance.

\subsection{Weight-space LoRA transfer and merging}

Weight-space methods address interference or compatibility between parameter updates. TIES-Merging trims small task-vector entries, resolves sign conflicts, and merges sign-consistent parameters~\citep{yadav2023ties}. KnOTS specifically addresses LoRA merging by jointly aligning updates through SVD before applying a merging rule~\citep{stoica2024knots}. These methods primarily combine capabilities from multiple adapted models. DART instead transfers one existing adapter to a distilled target; it does not aggregate a library of task adapters.

CASA is the closest evaluated baseline in the video reuse setting. It projects LoRA and model-drift updates into source singular coordinates and arbitrates routing interference within spectral clusters~\citep{wang2026casa}. DART shares the concern for parameter compatibility, but uses transported rank-one channels as the directions for subsequent response calibration. Static compatibility is consequently an input to its construction, while measured incremental denoiser responses under the target schedule determine the channel coefficients. Our CASA comparison evaluates complete transfer procedures; it does not isolate which coordinate transformation is superior.

\subsection{Adapter reparameterization and mapping}

Reparameterization changes an adapter's coordinates or factors, while mapping can connect different model representations. LoRA-X constrains the source adapter to a base-model subspace and transfers it to compatible target subspaces without additional transfer training~\citep{farhadzadeh2025lorax}. This is prior evidence that training-free cross-model transfer is possible; DART does not claim that idea itself. DART starts from an existing LoRA update and additionally measures how its transported channels act under the target denoising configuration. Preserving rank or spectral energy through a coordinate change does not establish preservation of the generated effect.

X-Adapter pursues plugin compatibility through trained feature-mapping layers between old and upgraded diffusion models~\citep{ran2024xadapter}. It keeps the old model's plugin interface and uses remapped features to guide the upgraded model. This differs from DART's conversion into a target-loadable LoRA with fixed channel coefficients, without learning an additional feature-mapping network. The comparison concerns the representation and conversion procedure, not an unmeasured performance ranking.

\subsection{Function-space alignment and response matching}

Function-space methods use model behavior as the transfer signal. Knowledge distillation matches teacher outputs~\citep{hinton2015distilling}, while FitNets also uses intermediate teacher representations to guide a student~\citep{romero2015fitnets}. Progressive distillation and consistency models connect related ideas to shorter diffusion trajectories~\citep{salimans2022progressive,song2023consistency}. DART operates after target distillation: its calibration descriptors concern the adapter-induced change relative to an unadapted baseline, rather than requiring the target to imitate the source model's complete output. Its local measured-response model also does not guarantee equality of the final generated videos.

Trans-LoRA uses synthetic task data to train transferred adapters for new base models~\citep{wang2024translora}. It illustrates why avoiding the original training dataset and avoiding further training are distinct properties. DART requires neither source training videos nor denoiser backpropagation during conversion, but it does use probe inputs and forward evaluations. Thus, source-data-free conversion should not be read as absence of calibration observations.

Gradient-free coefficient selection also has precedents. LoRAHub composes multiple task LoRAs using a few target-task examples and gradient-free optimization~\citep{huang2024lorahub}. DART instead fits coefficients for rank-one channels of one transported adapter using measured response descriptors across target timesteps and a regularized linear solve. The distinction lies in the coefficient granularity, measured signal, and deployment problem, rather than the absence of gradients alone.

\subsection{DART's specific combination and scope}

DART combines coordinate transport with \textbf{target-schedule-conditioned, channel-level, forward-only calibration for distilled image-to-video LoRA reuse}. Each qualifier specifies a concrete part of the method:
\begin{itemize}
\item \textbf{Target-schedule-conditioned:} probes and target timesteps enter the measured response operator. Each I2V probe specifies a latent state, text prompt, and reference image, held fixed within a paired response measurement. Conversion produces one fixed coefficient vector for that configuration, not a separate runtime coefficient vector at every step.
\item \textbf{Channel-level:} coefficients weight fixed rank-one directions within the transported update, rather than only scaling the complete adapter or weighting multiple task adapters.
\item \textbf{Forward-only:} paired finite-difference evaluations at the bridge estimate channel responses, followed by the closed-form regularized solve. \emph{Training-free} refers to the absence of iterative denoiser or LoRA training; calibration still fits coefficients and incurs forward-evaluation cost.
\item \textbf{Distilled I2V LoRA reuse:} the converted adapter is deployed on the existing few-step image-conditioned target. The purpose is to recover useful adapter effects or reduce negative transfer while evaluating image-conditioning fidelity separately from functional retention. The reported evaluations do not establish transfer performance for text-to-video generation.
\end{itemize}

Table~\ref{tab:related_work_boundaries} summarizes these conceptual boundaries. Our positioning is the specific combination of these choices, not a claim that coordinate alignment, response matching, or gradient-free fitting is individually new. The reported component ablations support complementary effects in the evaluated setting; they do not establish superiority over every work discussed here or recovery for every adapter.

\begin{table}[!htbp]
\centering
\small
\setlength{\tabcolsep}{4pt}
\caption{\textbf{Conceptual boundaries of DART.} Categories describe the primary object of a method, not mutually exclusive families or additional experimental comparisons.}
\label{tab:related_work_boundaries}
\begin{tabular}{@{}>{\raggedright\arraybackslash}p{0.23\linewidth} >{\raggedright\arraybackslash}p{0.33\linewidth} >{\raggedright\arraybackslash}p{0.39\linewidth}@{}}
\toprule
\textbf{Perspective} & \textbf{Primary object} & \textbf{DART's relationship} \\
\midrule
Weight-space transfer / merging & Parameter updates and their interference or compatibility & Transport supplies directions; target-schedule responses additionally determine their weights. \\
\addlinespace
Reparameterization / mapping & Adapter factors, coordinates, or model representations & Coordinate preservation alone does not guarantee preservation of the generated effect. \\
\addlinespace
Function-space / response matching & Model outputs or intermediate responses & Calibration fits descriptors of incremental adapter responses on fixed probes and target timesteps. \\
\addlinespace
DART combination & Transported rank-one channels and their measured responses & Forward-only conversion yields a fixed LoRA for the selected distilled I2V target. \\
\bottomrule
\end{tabular}
\end{table}
\FloatBarrier

\FloatBarrier
\clearpage
\section{LoRA Inventory}
\label{app:lora_inventory}

We initially collected eight public LoRAs for the high-noise branch of Wan2.2-I2V-A14B. They span animation and anime styles, subject actions, camera motion, rotation, and local gestures. The primary benchmark retains LoRAs 01, 03, 04, 05, 06, and 07. LoRAs 02 and 08 are excluded because their source-side functional gains are close to zero, making the denominator of the ratio-based FGR unstable. The exclusion is determined from the source anchors before target outcomes are inspected.

Table~\ref{tab:lora_inventory} records each adapter's identity, benchmark status, and functional role. The adapter IDs are fixed throughout the paper and match the generation manifest. Reporting all eight initially collected adapters makes the two exclusions and the retained benchmark explicit.

\begin{table}[H]
\centering
\footnotesize
\setlength{\tabcolsep}{5pt}
\caption{\textbf{Source LoRA inventory.} Status indicates inclusion in the six-adapter primary benchmark. Functional role summarizes the effect targeted by each source adapter.}
\label{tab:lora_inventory}
\begin{tabular}{@{}l l >{\raggedright\arraybackslash}p{1.55in} >{\raggedright\arraybackslash}p{2.75in}@{}}
\toprule
\textbf{ID} & \textbf{Status} & \textbf{Name} & \textbf{Functional role} \\
\midrule
01 & Retained & City the Animation & Elastic 2D animation and secondary motion \\
02 & Excluded & KungFu & Continuous martial-arts actions \\
03 & Retained & Fly & Takeoff, hovering, diving, and fast flight \\
04 & Retained & ArcShot & Arc or orbit camera motion with parallax \\
05 & Retained & Anime Lore & Cel-animation appearance and anime lighting \\
06 & Retained & DroneShot & Rising, retreating, overhead, descending, and wide-area aerial camera motion \\
07 & Retained & Turntable 360 & Subject rotation with a mostly fixed camera \\
08 & Excluded & Paw Pose & Paw raising, waving, joining, and reaching gestures \\
\bottomrule
\end{tabular}
\end{table}

\begin{figure}[H]
\centering
\includegraphics[width=\linewidth]{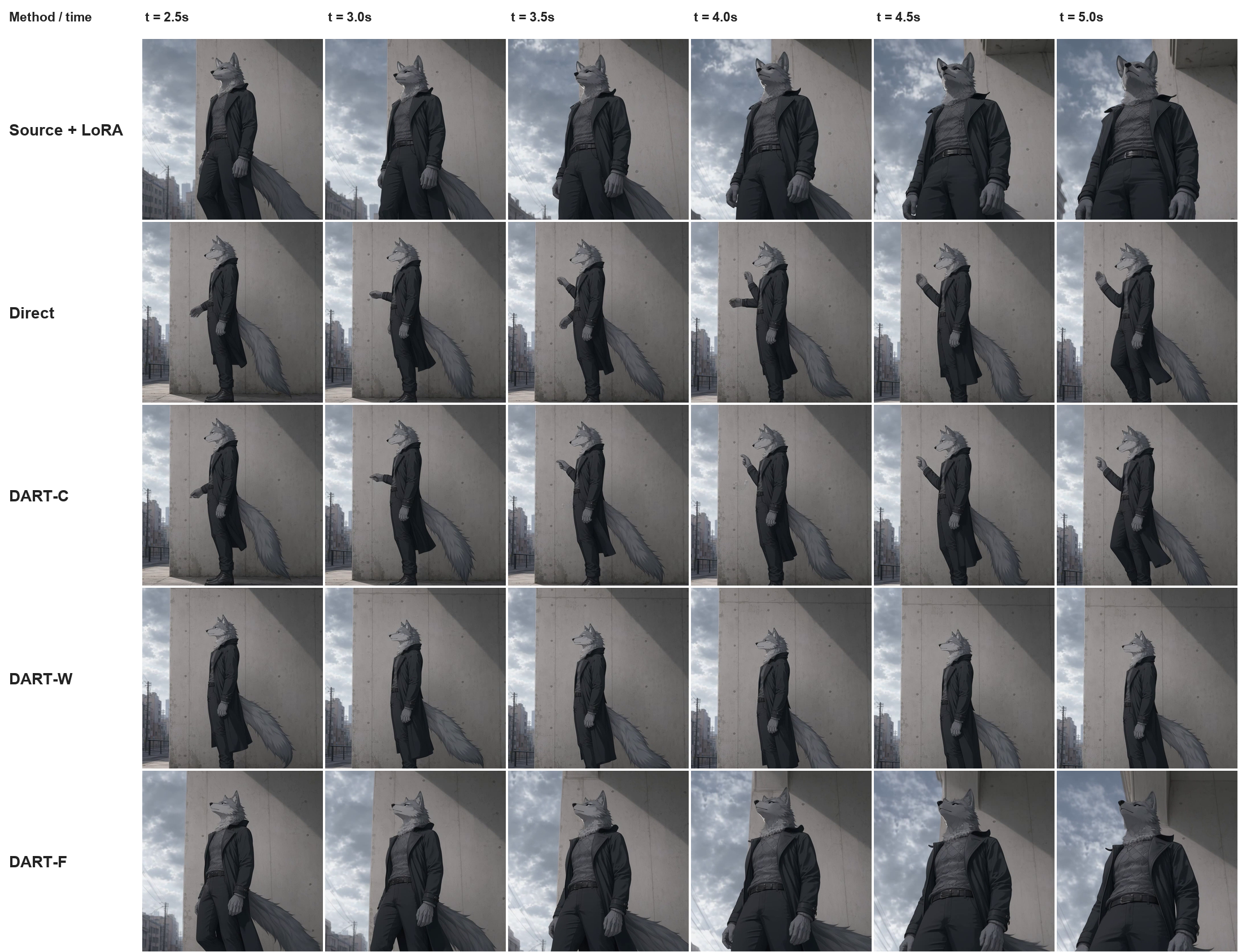}
\caption{\textbf{LoRA 04 (ArcShot) qualitative component comparison.} The calibration-only DART-C variant does not recover the requested camera motion, whereas DART-W recovers part of the arc and the full two-stage DART-F gives the strongest functional result.}
\label{fig:lora04_arcshot_case}
\end{figure}

\paragraph{LoRA 04 qualitative component case.}
LoRA 04 (ArcShot) targets an arc or orbit camera movement with parallax. The prompt used in Figure~\ref{fig:lora04_arcshot_case} is: \emph{arc shot, the camera arcs low around the subject in the first frame looking upward for a heroic angle, the subject holding a confident pose while clouds streak overhead.} Direct and DART-C largely preserve the side view and do not recover the requested low-angle arc. DART-W recovers part of the camera movement, while DART-F most clearly approaches the rising low-angle view exhibited by Source with the original LoRA. The visual ordering agrees with the primary FGR values reported in Appendix Table~\ref{tab:app_fgr_detail}: $-0.8459$ for Direct, $+0.3910$ for CASA, $+0.4910$ for DART-W, $-0.8104$ for DART-C, and $+0.5377$ for DART-F.

\FloatBarrier

\FloatBarrier
\clearpage
\section{Additional Exploratory Diagnostics}
\label{app:exploration_fig}

The static geometry and source-to-target response comparison appear in Figure~\ref{fig:geometry_response} in Section~\ref{sec:exploration}. This appendix provides the supporting four-anchor summary and global-scale sweep. These diagnostics are descriptive and should be interpreted alongside adapter-level quality and functional outcomes.

\begin{figure*}[!htbp]
\centering
\includegraphics[width=0.58\textwidth,trim=0 0 0 22bp,clip]{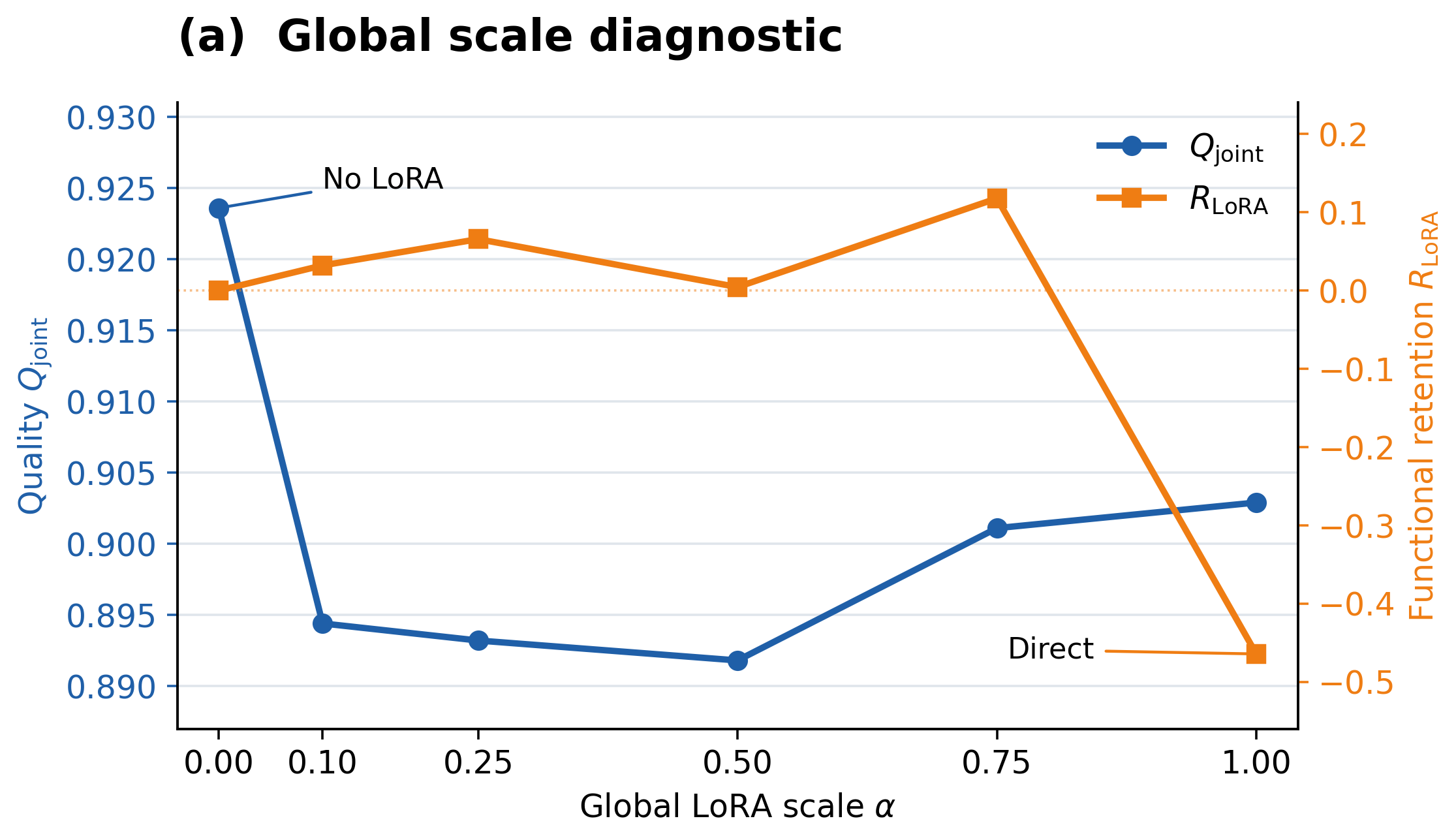}
\caption{\textbf{Global-scale transfer diagnostic.} Joint quality and functional retention under a shared global adapter scale, before DART. Table~\ref{tab:app_scale} reports all tested scales. The static-geometry and target-schedule response diagnostics are shown in the main text (Figure~\ref{fig:geometry_response}).}
\label{fig:scale_sweep}
\end{figure*}

\FloatBarrier

\subsection{Four-anchor and static-geometry summary}

\begin{table*}[!ht]
\centering
\tablefont
\caption{\textbf{Four-anchor and static-geometry summary.} $Q_{\rm joint}$ is the mean of VBench-Q and I2V-Avg. $Q_C$ and $Q_D$ denote target no-LoRA and target Direct joint quality, respectively, and $\Delta Q=Q_D-Q_C$. FGR uses source-selected CSD for LoRA 01 and VideoMAE for LoRA 03--07. Coordinate defect is reported in units of $10^{-3}$. The best, second-best, and third-best values in each directional metric column are highlighted with \colorbox{RankFirst}{green}, \colorbox{RankSecond}{yellow}, and \colorbox{RankThird}{blue} backgrounds, respectively.}
\label{tab:exploration_anchor_geometry}
\resizebox{\textwidth}{!}{%
\begin{tabular*}{\textwidth}{@{\extracolsep{\fill}}l c c c c c@{}}
\toprule
\textbf{Adapter} & $Q_C\uparrow$ & $Q_D\uparrow$ & $\Delta Q\uparrow$ & \textbf{Direct FGR}$\uparrow$ & \textbf{Coord. defect}$\downarrow$ \\
\midrule
LoRA 01 & \thirdresult{0.9249} & 0.8045 & $-0.1204$ & $-1.2714$ & \firstresult{2.635} \\
LoRA 03 & 0.9231 & 0.9162 & $-0.0070$ & \thirdresult{$-0.5381$} & 2.652 \\
LoRA 04 & \firstresult{0.9356} & \firstresult{0.9329} & \thirdresult{-0.0027} & $-0.8459$ & \secondresult{2.647} \\
LoRA 05 & 0.9152 & 0.9012 & $-0.0140$ & \secondresult{+0.3004} & \thirdresult{2.648} \\
LoRA 06 & 0.9096 & \thirdresult{0.9301} & \firstresult{+0.0205} & $-0.7712$ & 2.652 \\
LoRA 07 & \secondresult{0.9333} & \secondresult{0.9326} & \secondresult{-0.0007} & \firstresult{+0.3397} & 2.652 \\
\midrule
\textbf{Mean or range} & 0.9236 & 0.9029 & $-0.0207$ & $-0.4644$ & 0.017 \\
\bottomrule
\end{tabular*}%
}
\end{table*}

\FloatBarrier
\subsection{Global scale sweep}

\begin{table}[H]
\centering
\tablefont
\caption{\textbf{Global scale sweep.} All rows use the same cases, seeds, and scoring protocol. Functional retention is the adapter-macro $R_{\rm LoRA}$ in Equation~\ref{eq:r_lora}. The $\alpha=0$ row is target no-LoRA from Table~\ref{tab:app_anchor}, with zero recovery by definition; $\alpha=1$ matches Direct in Table~\ref{tab:main}. The best, second-best, and third-best entries in each metric column are highlighted with \colorbox{RankFirst}{green}, \colorbox{RankSecond}{yellow}, and \colorbox{RankThird}{blue} backgrounds.}
\label{tab:app_scale}
\resizebox{\textwidth}{!}{%
\begin{tabular}{@{}>{\centering\arraybackslash}p{1.35in} >{\centering\arraybackslash}p{2.075in} >{\centering\arraybackslash}p{2.075in}@{}}
\toprule
$\boldsymbol{\alpha}$ & \textbf{Mean $Q_{\rm joint}$}$\uparrow$ & $R_{\rm LoRA}\uparrow$ \\
\midrule
0.00 & \firstresult{0.9236} & 0.0000 \\
0.10 & 0.8944 & \thirdresult{0.0321} \\
0.25 & 0.8932 & \secondresult{0.0657} \\
0.50 & 0.8918 & 0.0044 \\
0.75 & \thirdresult{0.9011} & \firstresult{0.1178} \\
1.00 & \secondresult{0.9029} & $-0.4644$ \\
\bottomrule
\end{tabular}%
}
\end{table}

\FloatBarrier

\FloatBarrier
\clearpage
\section{Theory and Notation}
\label{app:theory}

This appendix supplies the definitions and derivations referenced by the six main method equations. It distinguishes properties of coordinate transport, measured linear responses, and the nonlinear denoiser. Each claim is stated with the conditions needed for its proof.

\subsection{Probe domain and response space}
\label{app:probe_domain}

Let an adapted layer have input and output dimensions $n_{\rm in}$ and $n_{\rm out}$. Its weights and updates belong to $\mathbb R^{n_{\rm out}\times n_{\rm in}}$. For each fixed configuration of the remaining layers, the denoiser output belongs to a finite dimensional space $\mathcal Y=\mathbb R^q$. A probe contains a latent state, a prompt, and a reference image. Their domains are $\mathcal Z$, $\mathcal P$, and $\mathcal I$. The probe domain, target time interval, target schedule, and response space are
\begin{equation}
\begin{aligned}
\Omega&=\mathcal Z\times\mathcal P\times\mathcal I,
&\tau_j&\in[0,T],\quad T>0,\\
S_t&=(\tau_1,\ldots,\tau_m),
&\mathcal H_t&=L^2(\Omega,\mu;\mathcal Y)^m.
\end{aligned}
\label{eq:domain_main}
\end{equation}
The measure $\mu$ is a probability measure on probes. The empirical analysis uses finitely many probes $x_n$ with masses $\mu_n>0$ that sum to one. Step weights $\omega_j>0$ also sum to one. For square integrable responses, the inner product is
\begin{equation}
\langle u,v\rangle_{\mathcal H_t}
=\sum_{j=1}^{m}\omega_j\int_\Omega u_j(x)^\top v_j(x)\,d\mu(x).
\label{eq:app_inner}
\end{equation}
All matrix derivatives below hold the probe, target time, and remaining layer weights fixed. Let $f_{t,j}(W;x)$ be the target denoiser output at time $\tau_j$. The exact response based at the bridge is
\begin{equation}
\mathcal R_t(A)
=\bigl(f_{t,j}(W_b+A;x)-f_{t,j}(W_b;x)\bigr)_{j=1}^{m}
\in\mathcal H_t.
\label{eq:app_response}
\end{equation}
Both terms use the same baseline construction. Changing the baseline would change the response being compared.

\subsection{Transport construction and invariants}
\label{app:transport_construction}

Write $\Delta=W_t-W_s$, $K=\Pi_{r_b}(\Delta)$, $W_b=W_s+K$, and $R=\Delta-K$. Choose source and bridge singular subspace bases $U_s,U_b$ with the same number of orthonormal columns, and similarly choose $V_s,V_b$. The cluster partition $\mathcal G$ groups nearly equal singular values. Block orthogonal Procrustes rotations $Q_U$ and $Q_V$ align corresponding clusters. Their dimensions match the selected left and right bases, respectively. The maps are
\begin{equation}
P_U=U_bQ_UU_s^\top,\qquad
P_V=V_bQ_VV_s^\top,\qquad
\widetilde C=P_UCP_V^\top.
\label{eq:app_transport_maps}
\end{equation}
For a nonzero update, the static diagnostic compares its coordinates under the source and bridge bases
\begin{equation}
\delta_{\rm coord}(C)
=\frac{1}{\|C\|_F}
\min_{Q_U,Q_V\in\mathbb O_{\mathcal G}}
\left\|U_b^\top CV_b-Q_U(U_s^\top CV_s)Q_V^\top\right\|_F.
\label{eq:app_defect}
\end{equation}
Here $\mathbb O_{\mathcal G}$ denotes the appropriate block orthogonal groups. This diagnostic is separate from the functional response.

Compact subspace maps need not be orthogonal on the entire ambient space. Proposition~\ref{prop:invariants} requires that they be isometries on the spaces occupied by the adapter. For the construction above, sufficient support conditions are
\begin{equation}
U_sU_s^\top C=C,\qquad CV_sV_s^\top=C.
\label{eq:app_transport_support}
\end{equation}
To prove the proposition, write a compact singular value decomposition $C=L\Sigma H^\top$. The support conditions imply that $P_UL$ and $P_VH$ have orthonormal columns. Consequently,
\begin{equation}
\widetilde C=(P_UL)\Sigma(P_VH)^\top,\qquad
\widetilde C^\top\widetilde C
=(P_VH)\Sigma^2(P_VH)^\top.
\label{eq:app_transport_spectrum}
\end{equation}
This is a singular value decomposition with the same nonzero singular values. Therefore,
\begin{equation}
\rank(\widetilde C)=\rank(C),\qquad
\|\widetilde C\|_F^2=\operatorname{tr}(\Sigma^2)=\|C\|_F^2.
\label{eq:app_transport_norm}
\end{equation}
If the support conditions fail, the maps can remove components outside the selected spaces, and the invariant claim does not apply. In plain terms, coordinate transport preserves adapter size only when its subspaces contain the adapter.

\subsection{Measured channels and visible equivalence}
\label{app:probe_operator}

Let $C_k$ be the fixed rank-one channels with $\widetilde C=\sum_{k=1}^{r_c}C_k$. Paired forward evaluations define
\begin{equation}
\widehat g_{d,j,k}(x)
=\frac{f_{t,j}(W_b+\epsilon C_k;x)-f_{t,j}(W_b-\epsilon C_k;x)}
{2\epsilon},\qquad \epsilon>0.
\label{eq:fd_main}
\end{equation}
The target index $d$ includes the selected schedule, probes, and response construction. With $\widehat g_{d,k}=(\widehat g_{d,j,k})_{j=1}^{m}$, the linear measured operator and its visible rank are
\begin{equation}
\widehat{\mathcal G}_d\colon\mathbb R^{r_c}\rightarrow\mathcal H_t,\qquad
\widehat{\mathcal G}_da=\sum_{k=1}^{r_c}a_k\widehat g_{d,k},
\qquad r_{\rm vis}=\rank(\widehat{\mathcal G}_d).
\label{eq:operator_main}
\end{equation}
This operator is linear by construction. It is a local approximation to the nonlinear denoiser response, not an assertion that the full video generator is linear in $a$.

Two coefficient vectors have the same measured response exactly when
\begin{equation}
a\sim_d a'
\quad\Longleftrightarrow\quad
\widehat{\mathcal G}_d(a-a')=0
\quad\Longleftrightarrow\quad
a-a'\in\Null(\widehat{\mathcal G}_d).
\label{eq:equivalence_main}
\end{equation}
The equality follows directly from linearity. Rank nullity then gives
\begin{equation}
r_c=r_{\rm vis}+\dim\Null(\widehat{\mathcal G}_d).
\label{eq:app_rank_nullity}
\end{equation}
In plain terms, the target probes determine how many distinct coefficient changes can be identified.

\subsection{Descriptor construction and the coefficient solution}
\label{app:calibration_construction}

For the empirical probe measure, let $E$ stack response entries with weights $\sqrt{\omega_j\mu_n}$. Then $E$ preserves the empirical response norm and the finite matrix $G_d=E\widehat{\mathcal G}_d$ has one column per channel. For a response $g$, define its step mean $\bar g_n=\sum_j\omega_jg_j(x_n)$. Let the matrices $\mathcal A$ and $\mathcal C$ act on this weighted representation by
\begin{equation}
\mathcal A(Eg)=\bigl(\sqrt{\mu_n}\bar g_n\bigr)_n,\qquad
\mathcal C(Eg)=\bigl(\sqrt{\omega_j\mu_n}(g_j(x_n)-\bar g_n)\bigr)_{j,n}.
\label{eq:app_descriptors}
\end{equation}
Thus $\mathcal A$ forms the step mean, while $\mathcal C$ retains deviations from that mean.

Let $e_s\neq0$ be the source mean response in the same descriptor space under the fixed probe correspondence, and set $u_s=e_s/\|e_s\|_2$. The source and target descriptors must have matching dimensions. The desired relative amplitude is $\rho\geq0$, and $P_s^\perp=I-u_su_s^\top$ removes the source direction. With $\lambda_{\rm mag},\lambda_{\rm step}\geq0$ and $\lambda_{\rm id}>0$, the expanded objective is
\begin{equation}
\begin{aligned}
\mathcal L_d(a)={}&
\|P_s^\perp\mathcal A G_da\|_2^2\\
&+\lambda_{\rm mag}\bigl(u_s^\top\mathcal A G_da-\rho\|e_s\|_2\bigr)^2\\
&+\lambda_{\rm step}\|\mathcal C G_da\|_2^2
+\lambda_{\rm id}\|a-a_0\|_2^2.
\end{aligned}
\label{eq:app_objective_expanded}
\end{equation}
The first three terms are written as one least squares residual by defining
\begin{equation}
B_d=\begin{bmatrix}
P_s^\perp\mathcal A G_d\\
\sqrt{\lambda_{\rm mag}}u_s^\top\mathcal A G_d\\
\sqrt{\lambda_{\rm step}}\mathcal C G_d
\end{bmatrix},
\qquad
b_d=\begin{bmatrix}
0\\
\sqrt{\lambda_{\rm mag}}\rho\|e_s\|_2\\
0
\end{bmatrix}.
\label{eq:app_stacked_system}
\end{equation}
Each zero denotes a vector of the corresponding block dimension. The matrices $\mathcal A$, $\mathcal C$, and the source descriptors are fixed while optimizing $a$. Differentiating Equation~\ref{eq:objective_main} yields
\begin{equation}
(B_d^\top B_d+\lambda_{\rm id}I)a_d^\star
=B_d^\top b_d+\lambda_{\rm id}a_0.
\label{eq:app_ridge_normal}
\end{equation}
For every nonzero coefficient vector $z$,
\begin{equation}
z^\top(B_d^\top B_d+\lambda_{\rm id}I)z
=\|B_dz\|_2^2+\lambda_{\rm id}\|z\|_2^2>0.
\label{eq:app_positive_definite}
\end{equation}
The Hessian is twice this positive definite matrix, so the objective has a unique minimizer and Equation~\ref{eq:closed_form_main} follows. In plain terms, the identity penalty makes calibration well defined even when the measured channels are redundant.

\subsection{Global scaling and invisible coefficients}

Define $v_d=B_da_0$ and $y_d=b_d$. A single scale produces only descriptors of the form $\alpha v_d$. Orthogonal projection gives
\begin{equation}
\min_{\alpha\in\mathbb R}\|\alpha v_d-y_d\|_2^2
=\|P_{\operatorname{span}(v_d)}^\perp y_d\|_2^2.
\label{eq:scale_limit_main}
\end{equation}
When $v_d\neq0$, differentiating the scalar quadratic gives
\begin{equation}
\alpha^\star=\frac{v_d^\top y_d}{v_d^\top v_d},\qquad
\|\alpha^\star v_d-y_d\|_2^2
=\|y_d\|_2^2-\frac{(v_d^\top y_d)^2}{\|v_d\|_2^2}.
\label{eq:app_scale_projection}
\end{equation}
If $v_d=0$, the error is $\|y_d\|_2^2$ for every scale and the same projection identity holds. The minimum is positive exactly when $y_d$ is outside the span of $v_d$. This proves the scalar part of Proposition~\ref{prop:calibration_main}. In plain terms, scaling cannot reach a descriptor direction outside the one produced by the original coefficient vector.

Now let $N=\Null(\widehat{\mathcal G}_d)$ and decompose $a-a_0=a_\perp+a_N$ with $a_N\in N$ and $a_\perp\perp N$. Since $B_d$ is a composition of measured responses and fixed linear descriptors, $B_da_N=0$. Therefore,
\begin{equation}
\mathcal L_d(a_0+a_\perp+a_N)
=\|B_d(a_0+a_\perp)-b_d\|_2^2
+\lambda_{\rm id}\|a_\perp\|_2^2
+\lambda_{\rm id}\|a_N\|_2^2.
\label{eq:app_null_penalty}
\end{equation}
For fixed $a_\perp$, the unique minimizing $a_N$ is zero. Thus
\begin{equation}
P_{\Null(\widehat{\mathcal G}_d)}(a_d^\star-a_0)=0.
\label{eq:null_main}
\end{equation}
The same argument applies to $\Null(B_d)$, which can be larger if descriptors discard measured information. In plain terms, calibration leaves unidentifiable coefficient combinations at identity.

\subsection{Calibration bias}

The bias relation does not require $B_d$ to have full column rank. Let $B_d=U\Sigma V^\top$ be its compact singular value decomposition with positive singular values $\sigma_\ell$, and let $r_d=b_d-B_da_0$. Define the unregularized least squares solution closest to $a_0$ as $a_d^\dagger=a_0+B_d^\dagger r_d$, where the dagger on $B_d$ denotes its pseudoinverse. Solving the normal equation in singular coordinates gives
\begin{equation}
\begin{aligned}
a_d^\star-a_0
&=\sum_\ell\frac{\sigma_\ell}{\sigma_\ell^2+\lambda_{\rm id}}
(u_\ell^\top r_d)v_\ell,\\
a_d^\dagger-a_0
&=\sum_\ell\frac{u_\ell^\top r_d}{\sigma_\ell}v_\ell.
\end{aligned}
\label{eq:app_ridge_expansion}
\end{equation}
Subtracting the unregularized normal equation also gives
\begin{equation}
(B_d^\top B_d+\lambda_{\rm id}I)(a_d^\star-a_d^\dagger)
=-\lambda_{\rm id}(a_d^\dagger-a_0).
\label{eq:app_ridge_difference}
\end{equation}
Each identifiable correction is multiplied by $\sigma_\ell^2/(\sigma_\ell^2+\lambda_{\rm id})$. In plain terms, identity regularization suppresses corrections most strongly along weakly measured descriptor directions.

\subsection{Bridge to deployment response}

The calibrated response is measured at $W_b$, but deployment uses $W_t=W_b+R$. For a fixed update $A$, assume the denoiser is twice continuously differentiable on a neighborhood of all matrices $W_b+sR+tA$ with $(s,t)\in[0,1]^2$. Assume the resulting responses and mixed derivatives are integrable in the response norm. Define
\begin{equation}
\Phi_j(s,t;x)=f_{t,j}(W_b+sR+tA;x).
\label{eq:app_phi}
\end{equation}
The difference between the deployment and bridge responses is the alternating sum of its four corners. Applying the fundamental theorem of calculus in each variable gives
\begin{equation}
\begin{aligned}
&f_{t,j}(W_t+A;x)-f_{t,j}(W_t;x)
-\mathcal R_{t,j}(A;x)\\
&\qquad=\int_0^1\int_0^1
D_W^2 f_{t,j}(W_b+sR+tA;x)[R,A]\,dt\,ds.
\end{aligned}
\label{eq:app_mixed_integral}
\end{equation}
This identity depends on the selected bridge through $R$ and on calibration through $A=C_t(a_d^\star)$. It vanishes when either is zero, and can also vanish through the mixed interaction. In plain terms, the error from applying a bridge-calibrated update at the target depends on how the remaining drift interacts with that update.

\subsection{Finite difference error}

For each channel and probe, let $g(s)=f_{t,j}(W_b+sC_k;x)$. Assume $g$ is three times continuously differentiable on $[-\epsilon,\epsilon]$ and $\|g'''(s)\|_2\leq M_{jk}(x)$ there. The vector valued integral remainder gives
\begin{equation}
\begin{aligned}
\widehat g_{d,j,k}(x)-g'(0)
={}&\frac{1}{4\epsilon}\int_0^\epsilon(\epsilon-u)^2g'''(u)\,du\\
&+\frac{1}{4\epsilon}\int_{-\epsilon}^{0}(\epsilon+u)^2g'''(u)\,du.
\end{aligned}
\label{eq:app_taylor}
\end{equation}
Bounding the integrands and evaluating the scalar integrals yields
\begin{equation}
\|\widehat g_{d,j,k}(x)-g'(0)\|_2
\leq\frac{\epsilon^2}{6}M_{jk}(x).
\label{eq:app_fd_bound}
\end{equation}
If these bounds are square integrable, stacking them also bounds the channel operator error. In plain terms, the probing radius controls the local approximation error under the stated smoothness assumption. This conclusion concerns the response estimate and does not bound the error of a full generated trajectory.

\subsection{Target specific conversion regret}

For each target configuration $d$, write $\mathcal L_d(a)=a^\top Q_da-2q_d^\top a+c_d$, where $Q_d=B_d^\top B_d+\lambda_{\rm id}I\succ0$ and $q_d=B_d^\top b_d+\lambda_{\rm id}a_0$. Completing the square gives
\begin{equation}
\mathcal L_d(a)
=\mathcal L_d(a_d^\star)
+(a-a_d^\star)^\top Q_d(a-a_d^\star).
\label{eq:app_square}
\end{equation}
To compare shared conversions, assume a common coefficient dimension and a fixed channel correspondence across target configurations. For a finite collection of targets with positive weights $p_d$ summing to one, a shared coefficient vector therefore incurs
\begin{equation}
\sum_d p_d\bigl(\mathcal L_d(a)-\mathcal L_d(a_d^\star)\bigr)
=\sum_d p_d\|a-a_d^\star\|_{Q_d}^2\geq0,
\label{eq:app_target_regret}
\end{equation}
where $\|z\|_{Q_d}^2=z^\top Q_dz$. Equality holds exactly when $a$ equals every target optimum. The best shared vector solves
\begin{equation}
a_{\rm shared}^\star
=\left(\sum_d p_dQ_d\right)^{-1}
\sum_d p_dQ_da_d^\star.
\label{eq:app_shared_optimum}
\end{equation}
Thus different optima imply a strictly positive minimum shared cost. In plain terms, target specific conversion is justified when the measured target objectives favor different coefficients, rather than merely because the targets have different names.

\FloatBarrier
\clearpage
\section{First-frame Reference Images}
\label{app:i2v_reference_images}

\begin{figure}[!htbp]
\centering
\begin{minipage}[t]{0.47\linewidth}
\centering
\includegraphics[width=\linewidth]{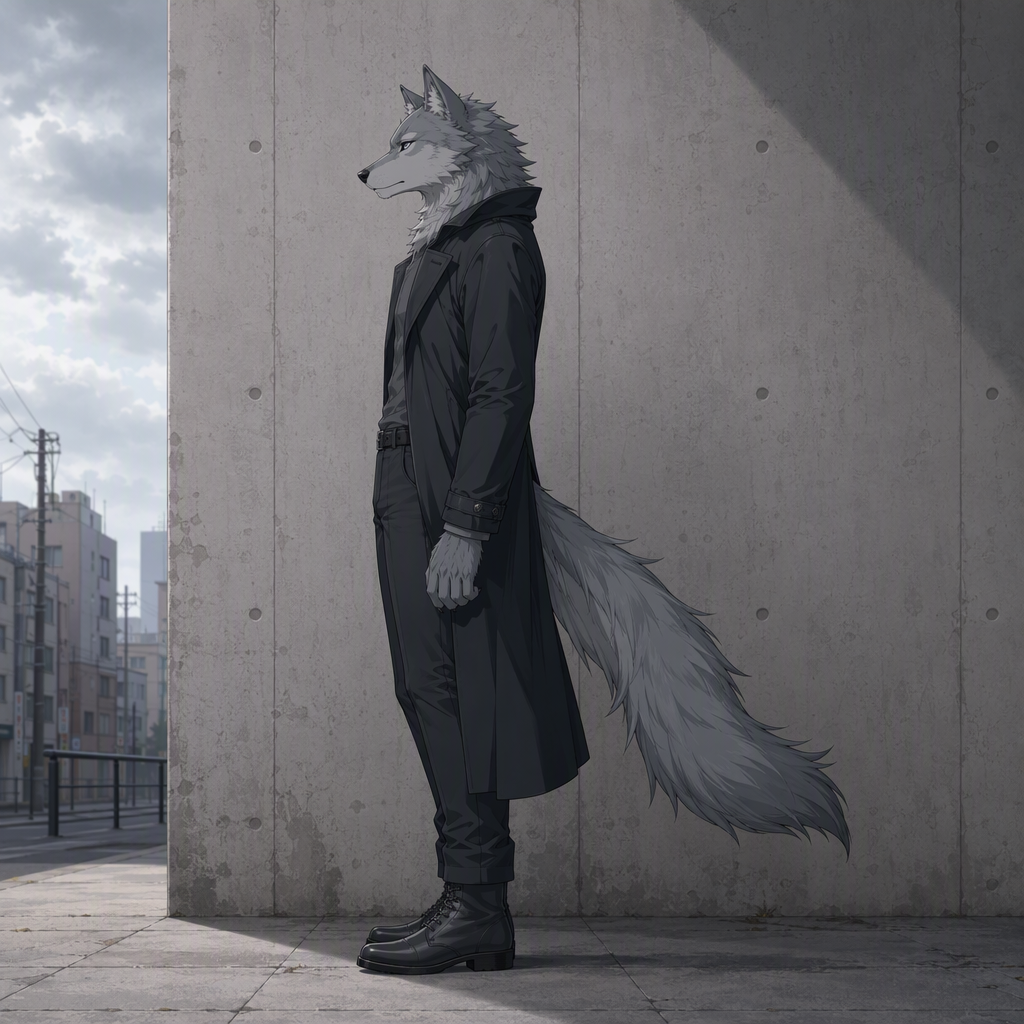}
\par\smallskip
{\small (a) Anime-style gray wolf, side view}
\end{minipage}
\hfill
\begin{minipage}[t]{0.47\linewidth}
\centering
\includegraphics[width=\linewidth]{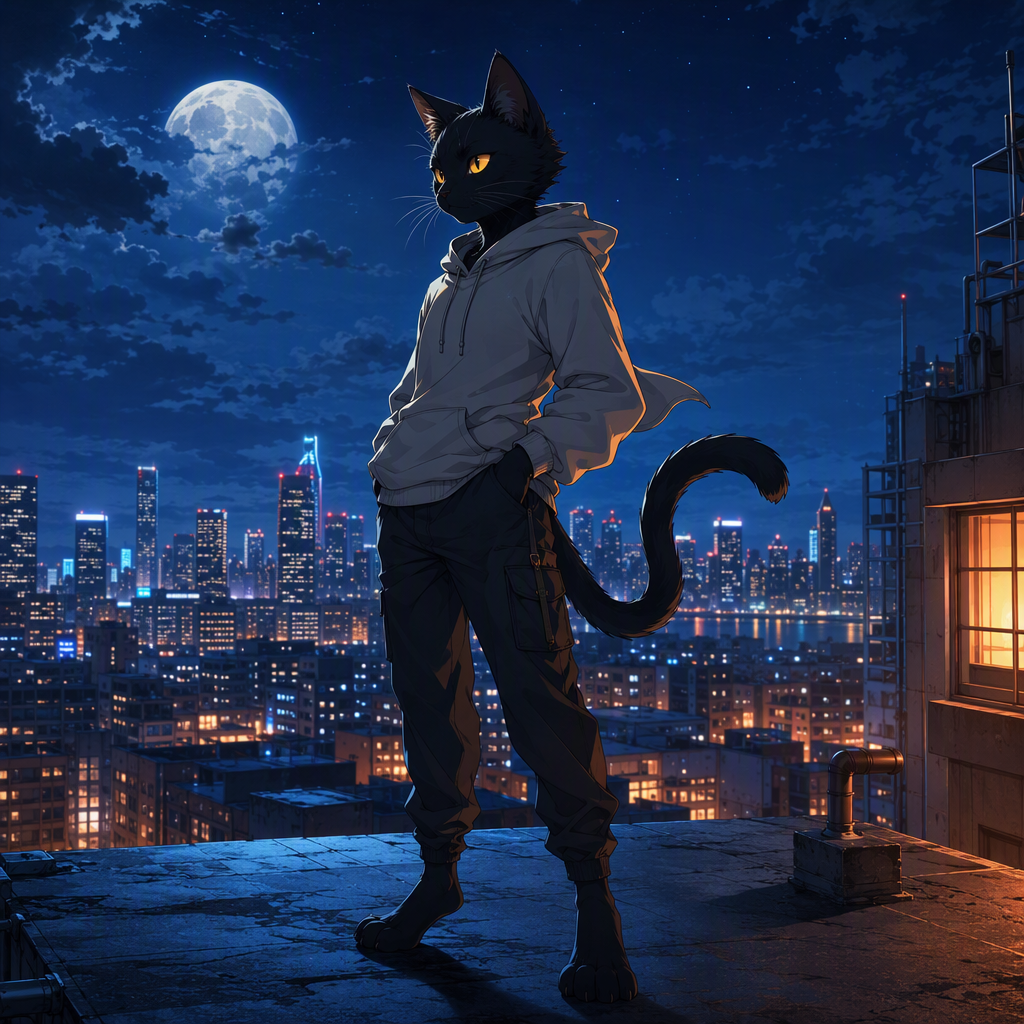}
\par\smallskip
{\small (b) Anime-style black cat in a nighttime scene}
\end{minipage}
\caption{\textbf{Examples of first-frame reference images for I2V generation.} Two of the eleven reference images used in the prompt--image combinations described in Section~\ref{sec:i2v_setup}. Images are displayed with their original aspect ratios.}
\label{fig:i2v_reference_images}
\end{figure}

The I2V generation design uses eleven first-frame reference images and eight prompts designed for each LoRA. Each prompt is paired with every reference image, giving $8\times11=88$ generated videos per adapter and method. Figure~\ref{fig:i2v_reference_images} displays two of the eleven input images: an anime-style gray wolf in side view and an anime-style black cat in a nighttime scene. These are conditioning inputs, rather than generated-video outputs.

\FloatBarrier

\FloatBarrier
\clearpage
\section{Probe Count and Subset Robustness}
\label{app:probe_robustness}
\renewcommand{\theHfigure}{probe.\arabic{figure}}
\renewcommand{\theHtable}{probe.\arabic{table}}
\renewcommand{\theHequation}{probe.\arabic{equation}}
\newcommand{\probeCountRows}{%
01 & 1 & 8 & 0.9854 & 27.35 & -- & -- & -- \\
01 & 2 & 28 & 0.9927 & 20.14 & 0.9973 & 7.99 & 18.53 \\
01 & 4 & 70 & 0.9969 & 11.30 & 0.9990 & 4.69 & 10.67 \\
01 & 6 & 28 & 0.9991 & 4.65 & 0.9995 & 3.06 & 5.58 \\
01 & 8 & 1 & 1.0000 & 0.00 & -- & -- & -- \\
\midrule
04 & 1 & 8 & 0.9762 & 22.00 & -- & -- & -- \\
04 & 2 & 28 & 0.9817 & 19.30 & 0.9789 & 21.72 & 39.65 \\
04 & 4 & 70 & 0.9863 & 16.59 & 0.9861 & 17.04 & 30.70 \\
04 & 6 & 28 & 0.9907 & 13.68 & 0.9898 & 14.35 & 22.73 \\
04 & 8 & 1 & 1.0000 & 0.00 & -- & -- & -- \\
\midrule
06 & 1 & 8 & 0.9671 & 27.55 & -- & -- & -- \\
06 & 2 & 28 & 0.9747 & 25.13 & 0.9801 & 21.84 & 44.24 \\
06 & 4 & 70 & 0.9849 & 18.87 & 0.9883 & 15.95 & 31.11 \\
06 & 6 & 28 & 0.9913 & 13.52 & 0.9925 & 12.45 & 17.71 \\
06 & 8 & 1 & 1.0000 & 0.00 & -- & -- & -- \\
}

\newcommand{\probeComplementRows}{%
01 & 0.8604 & 0.9998 & 0.9908 & 0.9980 & 0.9981 & 6.28 & 6.90 \\
04 & 0.4034 & 0.9976 & 0.9704 & 0.9720 & 0.9848 & 24.33 & 17.48 \\
06 & 0.5259 & 0.9969 & 0.9427 & 0.9827 & 0.9881 & 18.80 & 16.11 \\
}

\subsection{Controlled protocol and scope}
We examine whether response calibration is sensitive to the number and selection of probes, using LoRAs 01, 04, and 06 on the Wan2.2-I2V-A14B four-step high-noise checkpoint. We recover the first 32 spectral factors from each serialized DART-W adapter and calibrate 320 attention-projection modules through online LoRA branches. No additional bridge-weight delta is injected. This is a controlled diagnostic of the calibration procedure, rather than an exact reproduction of the historical bridge-based conversion coefficients or a new video-quality benchmark.

\paragraph{Inputs and numerical settings.}
A CPU random generator with seed 42 produces a shared pool of eight independent standard-normal latent tensors of shape $36\times3\times40\times60$. Text context is zero and no VAE-encoded reference image is supplied. An independently generated latent with seed 20260920 is held out from calibration. We evaluate scheduler step indices 0 and 1 of the high-noise branch, keeping the input fixed rather than advancing a denoising trajectory. Forward evaluations use BF16, cached responses use FP32, and the calibration system is solved in FP64. The central finite-difference radius is $\epsilon=0.2$. Direction and magnitude weights are both 1; step-balance and identity-centered regularization weights are 0.1 and 0.01, with an additional diagonal regularizer of $10^{-8}$.

\paragraph{Aggregation and held-out response.}
For subset $S$, channel responses and the full-adapter reference response are first averaged over probes; the resulting statistics define the calibration matrix $Q_S$, right-hand side $b_S$, and solution $a_S$. Averaging responses before constructing the objective differs from constructing and averaging per-probe objectives. We keep the former order fixed throughout this diagnostic. For the common held-out latent $x_h$, the reported actual response averages the two nonlinear denoiser differences:
\begin{equation}
r_S=\frac{1}{2}\sum_{t\in\{0,1\}}
\left[f_t(W_t+C(a_S);x_h)-f_t(W_t;x_h)\right].
\label{eq:probe_heldout_response}
\end{equation}
Every subset uses the same no-LoRA baseline. This directly measured response is distinct from a linear prediction obtained by combining cached channel derivatives. Let $r_8$ denote the fit using the entire eight-probe pool. Besides cosine similarity, we use two distinct vector distances:
\begin{equation}
d_8(r_S)=\frac{\|r_S-r_8\|_2}{\|r_8\|_2},
\qquad
d_{\rm sym}(x,y)=\frac{2\|x-y\|_2}{\|x\|_2+\|y\|_2}.
\label{eq:probe_distances}
\end{equation}
The first measures deviation from the finite-pool reference; the second compares two subsets symmetrically. Both include direction and magnitude changes. Neither is a pure magnitude error. The eight-probe fit is a reference estimate, not ground truth.

\paragraph{Subset coverage.}
The first round solves all $8+28+70+1=107$ subsets of sizes 1, 2, 4, and 8 per adapter. Actual held-out forward evaluations cover all eight singletons, four prespecified pairs, four prespecified four-probe subsets, and the full pool. The second round exhaustively evaluates all 28 two-probe, 70 four-probe, 28 six-probe, and one eight-probe subsets: 127 actual responses per adapter, or 381 in total. All 27 overlapping adapter--subset cases agree between rounds in the archived response-comparison metrics. We combine the 24 singleton measurements from the first round with the second round's results. Subsets share probes and one held-out input, so these counts do not represent independent test examples or random-seed replications.

\clearpage
\subsection{Probe-count results and the four-probe choice}

\begin{figure}[H]
\centering
\includegraphics[width=\textwidth]{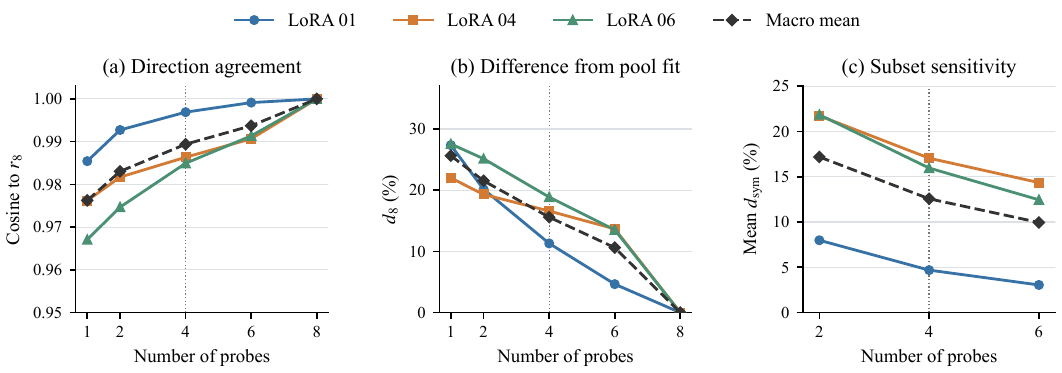}
\caption{\textbf{Held-out response stability versus probe count.} Panels (a,b) average all subsets against the eight-probe fit; (c) averages all equal-size subset pairs at sizes 2, 4, and 6. Dashed black curves are macro means; the vertical dotted line marks four probes. Eight-probe agreement is an identity comparison. All responses use one common synthetic held-out input.}
\label{fig:probe_count_stability}
\end{figure}

\begin{table}[H]
\centering
\small
\setlength{\tabcolsep}{4pt}
\caption{\textbf{Probe-count and subset sensitivity.} Means are over subsets or subset pairs. For 2, 4, and 6 probes there are 378, 2,415, and 378 pairs. Dashes mark unreported singleton pair statistics or the absence of distinct pairs at eight probes.}
\label{tab:probe_count_robustness}
\begin{tabular}{@{}ccrrrrrr@{}}
\toprule
& & & \multicolumn{2}{c}{Versus eight-probe fit} & \multicolumn{3}{c}{Within-count subset pairs}\\
\cmidrule(lr){4-5}\cmidrule(lr){6-8}
LoRA & Probes & Subsets & Cosine & $d_8$ (\%) & Cosine & \shortstack{Mean\\$d_{\rm sym}$ (\%)} & \shortstack{Max.\\$d_{\rm sym}$ (\%)}\\
\midrule
\probeCountRows
\bottomrule
\end{tabular}
\end{table}

\paragraph{Why four probes?}
The macro mean cosine to $r_8$ is 0.9762, 0.9830, 0.9894, and 0.9937 for 1, 2, 4, and 6 probes; corresponding $d_8$ values are 25.63\%, 21.52\%, 15.59\%, and 10.61\%. Four probes improve over both smaller budgets. The mean within-count $d_{\rm sym}$ falls from 17.18\% at two probes to 12.56\% at four and 9.95\% at six; the worst observed pair falls from 44.24\% to 31.11\% and 22.73\%. Thus four is an empirical compromise, with half the probe-dependent forward evaluations of eight under fixed channels and timesteps. This ratio concerns probing work, not total conversion time, which also includes setup and solving.

The improvements at six probes show no clear saturation at four. Moreover, more probes need not bring every adapter closer to the uncalibrated full-adapter response: its cosine with the calibrated response falls from 0.9642 to 0.9532 for LoRA 01 between two and eight probes, while increasing for LoRAs 04 and 06. That full-adapter response is itself not a functional ground truth on the distilled target.

\clearpage
\subsection{Changing the four-probe subset}

All 70 four-probe subsets form 35 disjoint complementary pairs. Table~\ref{tab:probe_complements} summarizes their intermediate quantities and two prespecified complementary pairs with actual nonlinear forward measurements: $A=(0123,4567)$ and $B=(0246,1357)$, using zero-based probe indices.

\begin{table}[H]
\centering
\small
\setlength{\tabcolsep}{3pt}
\caption{\textbf{Complementary four-probe subsets.} The first three cosine columns average all 35 complementary pairs. Actual cosine and symmetric vector distance use the two prespecified pairs $A$ and $B$. Linear and actual responses are both evaluated on the common held-out latent.}
\label{tab:probe_complements}
\begin{tabular}{@{}crrrrrrr@{}}
\toprule
& \multicolumn{3}{c}{Mean cosine, 35 pairs} & \multicolumn{2}{c}{Actual cosine} & \multicolumn{2}{c}{Actual $d_{\rm sym}$ (\%)}\\
\cmidrule(lr){2-4}\cmidrule(lr){5-6}\cmidrule(lr){7-8}
LoRA & Coefficients & Energy & Linear & $A$ & $B$ & $A$ & $B$\\
\midrule
\probeComplementRows
\bottomrule
\end{tabular}
\end{table}

Coefficient directions vary substantially for LoRAs 04 and 06, whereas channel-energy profiles and actual response directions are more consistent. Across all equal-size four-probe pairs, actual response cosine averages 0.9911 across adapters (Table~\ref{tab:probe_count_robustness}). This supports evaluating the resulting model response rather than inferring response instability from coefficients alone. However, the complementary-pair vector distances reach 24.33\%, and the exhaustive four-probe comparison reaches 31.11\%; subset effects are not negligible.

\paragraph{Adapter identity versus subset choice.}
For the balanced design of three adapters and the four prespecified four-probe subsets, a descriptive crossed sum-of-squares decomposition attributes 99.736\% of actual response-vector variation to adapter identity, 0.089\% to subset identity, and 0.175\% to their interaction. After unit-normalizing each response, the respective shares are 98.929\%, 0.367\%, and 0.704\%. Adapter differences dominate in this design even after removing response scale. These are descriptive shares for selected adapters and overlapping subsets on one input, not population variance estimates, significance tests, or causal contributions.

\subsection{Numerical checks and limits of the evidence}
\label{app:probe_numerics}

Identical baseline forwards are deterministic in the repeated-input check. Spectral-factor recovery covers 320 modules per adapter; for LoRA 01 the maximum relative truncation-tail norm is $2.37\times10^{-7}$, with no ordering violations among the first 32 factors. Nevertheless, stable responses do not imply stable calibration coefficients. Matrix conditioning can amplify changes in the probe statistics. Table~\ref{tab:probe_numerics} also shows sensitivity to the finite-difference radius, which prevents a claim of derivative convergence.

\begin{table}[H]
\centering
\small
\setlength{\tabcolsep}{5pt}
\caption{\textbf{Numerical diagnostics.} The first two columns give medians over four-probe subsets. The last two compare derivative directions with $\epsilon=0.2$, averaged over the first probe, two steps, and channels 0, 16, and 31. Values are approximate.}
\label{tab:probe_numerics}
\begin{tabular}{@{}crrrr@{}}
\toprule
LoRA & \shortstack{Relative $Q$ change\\versus eight probes} & \shortstack{Condition\\number} & \shortstack{Derivative cosine\\$\epsilon=0.1$} & \shortstack{Derivative cosine\\$\epsilon=0.4$}\\
\midrule
01 & 0.349 & 12,328 & 0.48 & 0.54\\
04 & 0.861 & 182 & 0.26 & 0.27\\
06 & 0.799 & 71 & 0.29 & 0.33\\
\bottomrule
\end{tabular}
\end{table}

The $\epsilon=0.001$ check has larger relative errors. BF16 precision and finite-radius nonlinearity are possible contributors, but this experiment does not separate them. The main stability measurements use actual nonlinear forwards, while their coefficients still depend on this finite-difference estimator.

The independent held-out latent tests behavior beyond the calibration inputs, but one synthetic input cannot rule out broader calibration overfitting. Without real text or reference-image variation, these results do not establish prompt/reference-image robustness, full-trajectory generalization, or stability of $Q_{\rm joint}$ and FGR. Four probes remain an empirical budget choice for the measured protocol.

\FloatBarrier
\clearpage
\section{Detailed Results and Diagnostics}
\label{app:details}

The following tables retain adapter-level summaries and diagnostic inventories. Primary quality and FGR use all 88 videos per adapter and method, covering the eight prompts crossed with eleven first-frame reference images. The method, adapter, and Direct anchor tables use the same cases; source and no-LoRA anchors use the corresponding 88-case evaluations. Each video uses the six-dimension quality mean, excluding dynamic degree, and the two-dimension I2V mean. Means and differences are calculated before rounding, so differences of displayed entries can differ in the final decimal place.

\subsection{Adapter-level quality and influence of LoRA 01}
\label{app:adapter_heterogeneity}

Table~\ref{tab:app_adapter_quality} reports all six adapters. DART-F improves $Q_{\rm joint}$ by $+0.1213$, $+0.0021$, $+0.0052$, and $+0.0160$ for LoRAs 01, 03, 04, and 05; the changes for LoRAs 06 and 07 are $-0.0255$ and $-0.0006$. Excluding LoRA 01, the mean quality gain is $-0.0006$, compared with $+0.0198$ across all six adapters. The positive aggregate quality gain is therefore driven by the large improvement on LoRA 01, and should not be interpreted as a uniform adapter-level benefit. This exclusion summary is descriptive and does not estimate uncertainty across seeds.

\begin{table}[H]
\centering
\tablefont
\caption{\textbf{Per-adapter quality and reference similarity.} $Q_{\rm joint}$ is the mean of VBench-Q and I2V-Avg, both normalized to a zero-to-one scale. Gain is DART-F minus Direct. Ref-DINO uses DINOv2 features~\citep{oquab2023dinov2}; it is supplementary and is not included in $Q_{\rm joint}$. The best, second-best, and third-best entries in each numeric column are highlighted with \colorbox{RankFirst}{green}, \colorbox{RankSecond}{yellow}, and \colorbox{RankThird}{blue} backgrounds, respectively.}
\label{tab:app_adapter_quality}
\resizebox{\textwidth}{!}{%
\begin{tabular}{@{}l ccc ccc@{}}
\toprule
\textbf{Adapter} & \textbf{Direct $Q_{\rm joint}$}$\uparrow$ & \textbf{DART-F $Q_{\rm joint}$}$\uparrow$ & \textbf{$Q_{\rm joint}$ gain}$\uparrow$ & \textbf{Direct Ref-DINO}$\uparrow$ & \textbf{DART-F Ref-DINO}$\uparrow$ & \textbf{Gain}$\uparrow$ \\
\midrule
LoRA 01 & 0.8045 & \thirdresult{0.9258} & \firstresult{+0.1213} & 0.3931 & 0.8969 & \firstresult{+0.5038} \\
LoRA 03 & 0.9162 & 0.9183 & +0.0021 & 0.8680 & 0.8459 & $-0.0221$ \\
LoRA 04 & \firstresult{0.9329} & \firstresult{0.9381} & \thirdresult{+0.0052} & \thirdresult{0.9073} & \firstresult{0.9552} & \secondresult{+0.0479} \\
LoRA 05 & 0.9012 & 0.9172 & \secondresult{+0.0160} & 0.8562 & \thirdresult{0.9008} & \thirdresult{+0.0446} \\
LoRA 06 & \thirdresult{0.9301} & 0.9046 & $-0.0255$ & \secondresult{0.9122} & 0.8037 & $-0.1086$ \\
LoRA 07 & \secondresult{0.9326} & \secondresult{0.9319} & $-0.0006$ & \firstresult{0.9263} & \secondresult{0.9300} & +0.0037 \\
\midrule
\textbf{Mean} & 0.9029 & 0.9227 & +0.0198 & & & \\
\bottomrule
\end{tabular}%
}
\end{table}

\begin{table}[H]
\centering
\tablefont
\caption{\textbf{Four-anchor diagnostic.} A is Source without LoRA, B is Source with the original LoRA, C is Target without LoRA, and D is Target with Direct LoRA. Each $Q_{\rm joint}$ value is the mean of the corresponding VBench-Q and I2V-Avg values. FGR uses CSD for LoRA 01 and VideoMAE for LoRA 03--07. Higher is better, and the best, second-best, and third-best entries in each numeric column are highlighted with \colorbox{RankFirst}{green}, \colorbox{RankSecond}{yellow}, and \colorbox{RankThird}{blue} backgrounds, respectively.}
\label{tab:app_anchor}
\resizebox{\textwidth}{!}{%
\begin{tabular}{@{}l rrrrr rrrr r@{}}
\toprule
\textbf{Adapter} & $Q_{\rm joint,A}\uparrow$ & $Q_{\rm joint,B}\uparrow$ & $Q_{\rm joint,C}\uparrow$ & $Q_{\rm joint,D}\uparrow$ & $(Q_{\rm joint,D}-Q_{\rm joint,C})\uparrow$ & \textbf{I2V-A}$\uparrow$ & \textbf{I2V-B}$\uparrow$ & \textbf{I2V-C}$\uparrow$ & \textbf{I2V-D}$\uparrow$ & \textbf{Direct FGR}$\uparrow$ \\
\midrule
LoRA 01 & 0.8975 & 0.8457 & \thirdresult{0.9249} & 0.8045 & $-0.1204$ & 0.9425 & 0.9072 & 0.9730 & 0.8789 & $-1.2714$ \\
LoRA 03 & 0.9042 & 0.9073 & 0.9231 & 0.9162 & $-0.0070$ & 0.9555 & 0.9622 & \thirdresult{0.9744} & 0.9658 & \thirdresult{$-0.5381$} \\
LoRA 04 & \secondresult{0.9205} & \thirdresult{0.9226} & \firstresult{0.9356} & \firstresult{0.9329} & \thirdresult{-0.0027} & \secondresult{0.9761} & \firstresult{0.9834} & \firstresult{0.9850} & \firstresult{0.9845} & $-0.8459$ \\
LoRA 05 & \thirdresult{0.9044} & 0.8884 & 0.9152 & 0.9012 & $-0.0140$ & 0.9508 & 0.9327 & 0.9623 & 0.9454 & \secondresult{+0.3004} \\
LoRA 06 & 0.9015 & \secondresult{0.9247} & 0.9096 & \thirdresult{0.9301} & \firstresult{+0.0205} & \thirdresult{0.9703} & \thirdresult{0.9797} & 0.9730 & \secondresult{0.9808} & $-0.7712$ \\
LoRA 07 & \firstresult{0.9294} & \firstresult{0.9327} & \secondresult{0.9333} & \secondresult{0.9326} & \secondresult{-0.0007} & \firstresult{0.9783} & \secondresult{0.9819} & \secondresult{0.9824} & \thirdresult{0.9781} & \firstresult{+0.3397} \\
\midrule
\textbf{Mean} & 0.9096 & 0.9036 & 0.9236 & 0.9029 & -0.0207 & & & & & \\
\bottomrule
\end{tabular}%
}
\end{table}

\begin{table}[H]
\centering
\tablefont
\caption{\textbf{Primary FGR by adapter and method.} The primary proxy is selected from source-side gains before target outcomes are inspected. It is CSD for LoRA 01 and VideoMAE for LoRA 03--07. A negative entry means that the transferred adapter lowers the primary score relative to the target no-LoRA baseline. Higher is better; the best, second-best, and third-best entries in each method column are highlighted with \colorbox{RankFirst}{green}, \colorbox{RankSecond}{yellow}, and \colorbox{RankThird}{blue} backgrounds.}
\label{tab:app_fgr_detail}
\resizebox{\textwidth}{!}{%
\begin{tabular}{@{}l l rrrrr@{}}
\toprule
\textbf{Adapter} & \textbf{Proxy} & \multicolumn{5}{c}{\textbf{Primary FGR}$\uparrow$} \\
\cmidrule(lr){3-7}
& & \textbf{Direct} & \textbf{CASA} & \textbf{DART-W} & \textbf{DART-C} & \textbf{DART-F} \\
\midrule
LoRA 01 & CSD & $-1.2714$ & $-4.0210$ & $-3.9538$ & \thirdresult{$-0.0043$} & $-0.0187$ \\
LoRA 03 & VideoMAE & \thirdresult{$-0.5381$} & $-0.6924$ & $-0.5015$ & $-0.8025$ & $-0.2578$ \\
LoRA 04 & VideoMAE & $-0.8459$ & \secondresult{+0.3910} & \secondresult{+0.4910} & $-0.8104$ & \firstresult{+0.5377} \\
LoRA 05 & VideoMAE & \secondresult{+0.3004} & \firstresult{+0.4249} & \firstresult{+0.5474} & \secondresult{+0.3467} & \secondresult{+0.5229} \\
LoRA 06 & VideoMAE & $-0.7712$ & $-0.1317$ & $-0.1266$ & $-0.8760$ & $-0.2527$ \\
LoRA 07 & VideoMAE & \firstresult{+0.3397} & \thirdresult{+0.3866} & \thirdresult{+0.3355} & \firstresult{+0.3629} & \thirdresult{+0.2780} \\
\midrule
\textbf{Mean} & & \thirdresult{$-0.4644$} & $-0.6071$ & $-0.5347$ & \secondresult{$-0.2973$} & \firstresult{+0.1349} \\
\bottomrule
\end{tabular}%
}
\end{table}

\begin{table}[H]
\centering
\tablefont
\caption{\textbf{Static geometry inventory.} Each adapter is measured on ten representative layers. Coordinate defect is lower-is-better; alignment, FGR, and direct quality are higher-is-better. Layer count and update norm are descriptive inventory fields and are not ranked. The best, second-best, and third-best entries in each directional metric column are highlighted with \colorbox{RankFirst}{green}, \colorbox{RankSecond}{yellow}, and \colorbox{RankThird}{blue} backgrounds; ties share a rank.}
\label{tab:app_geometry}
\resizebox{\textwidth}{!}{%
\begin{tabular}{@{}l cccccccc@{}}
\toprule
\textbf{Adapter} & \textbf{Layers} & \textbf{Coord. defect}$\downarrow$ & \textbf{Subspace}$\uparrow$ & \textbf{Spectral}$\uparrow$ & $\|C\|_F$ & \textbf{Direct FGR}$\uparrow$ & \textbf{CASA FGR}$\uparrow$ & \textbf{Direct $Q_{\rm joint}$}$\uparrow$ \\
\midrule
LoRA 01 & 10 & \firstresult{$0.002635\mathbin{\pm}0.000041$} & \firstresult{0.999553} & \firstresult{0.999999658} & 52.2873 & $-1.2714$ & $-4.0210$ & 0.8045 \\
LoRA 03 & 10 & $0.002652\mathbin{\pm}0.000024$ & \firstresult{0.999553} & \firstresult{0.999999658} & 0.1162 & \thirdresult{$-0.5381$} & $-0.6924$ & 0.9162 \\
LoRA 04 & 10 & \secondresult{$0.002647\mathbin{\pm}0.000026$} & \firstresult{0.999553} & \firstresult{0.999999658} & 2.1596 & $-0.8459$ & \secondresult{+0.3910} & \firstresult{0.9329} \\
LoRA 05 & 10 & \thirdresult{$0.002648\mathbin{\pm}0.000028$} & \firstresult{0.999553} & \firstresult{0.999999658} & 0.1711 & \secondresult{+0.3004} & \firstresult{+0.4249} & 0.9012 \\
LoRA 06 & 10 & $0.002652\mathbin{\pm}0.000019$ & \firstresult{0.999553} & \firstresult{0.999999658} & 2.0664 & $-0.7712$ & $-0.1317$ & \thirdresult{0.9301} \\
LoRA 07 & 10 & $0.002652\mathbin{\pm}0.000024$ & \firstresult{0.999553} & \firstresult{0.999999658} & 0.0957 & \firstresult{+0.3397} & \thirdresult{+0.3866} & \secondresult{0.9326} \\
\bottomrule
\end{tabular}%
}
\end{table}

\begin{table}[H]
\centering
\tablefont
\caption{\textbf{Target-schedule response diagnostic.} Direction cosine and target-to-source magnitude ratio compare fixed-state responses. Magnitude ratio is ideally one; norms and counts are descriptive. The best, second-best, and third-best direction-cosine entries are highlighted with \colorbox{RankFirst}{green}, \colorbox{RankSecond}{yellow}, and \colorbox{RankThird}{blue} backgrounds; the subset sizes are descriptive rather than optimization scores.}
\label{tab:app_response}
\resizebox{\textwidth}{!}{%
\begin{tabular}{@{}l rrrrr@{}}
\toprule
\textbf{Subset} & $\boldsymbol{n}$ & \textbf{Direction cosine}$\uparrow$ & \textbf{Magnitude ratio} & \textbf{Target norm} & \textbf{Source norm} \\
\midrule
Branch mismatched & 3 & \thirdresult{0.0650} & 1.2912 & 68.70 & 95.19 \\
Branch matched & 9 & \firstresult{0.4529} & 0.9556 & 90.25 & 96.04 \\
All & 12 & \secondresult{0.3559} & 1.0395 & 84.86 & 95.83 \\
\bottomrule
\end{tabular}%
}
\end{table}

\begin{table}[H]
\centering
\tablefont
\caption{\textbf{Calibrated coefficient behavior.} Identity means coefficients equal to one. Near-zero counts use magnitude below $0.05$; near-one counts use distance from one below $0.05$. Deviation is distance from the identity vector divided by the square root of the channel count. $\Delta\mathrm{FGR}=\mathrm{FGR}_{\mathrm{DART-F}}-\mathrm{FGR}_{\mathrm{Direct}}$ reports the functional gain over Direct; positive values indicate improvement. The best, second-best, and third-best deviation values are highlighted with \colorbox{RankFirst}{green}, \colorbox{RankSecond}{yellow}, and \colorbox{RankThird}{blue} backgrounds.}
\label{tab:app_coeff}
\resizebox{\textwidth}{!}{%
\begin{tabular}{@{}l ccccccc@{}}
\toprule
\textbf{Adapter} & \textbf{Channels} & \textbf{Mean} & \textbf{Range} & \textbf{Deviation} & \textbf{Near zero} & \textbf{Near one} & $\boldsymbol{\Delta\mathrm{FGR}}\uparrow$ \\
\midrule
LoRA 01 & 32 & $-0.0004$ & $[-0.0227,0.0172]$ & 1.0005 & 32 & 0  & $+1.2527$ \\
LoRA 03 & 32 & $-0.0000$ & $[-0.0008,0.0006]$ & 1.0000 & 32 & 0  & $+0.2803$ \\
LoRA 04 & 32 & 0.4687 & $[-0.0012,1.0000]$ & \secondresult{0.7290} & 7 & 12  & $+1.3836$ \\
LoRA 05 & 32 & 0.4579 & $[-0.0014,1.0000]$ & \firstresult{0.722507} & 9 & 11  & $+0.2225$ \\
LoRA 06 & 32 & $-0.0002$ & $[-0.0032,0.0023]$ & 1.0002 & 32 & 0  & $+0.5185$ \\
LoRA 07 & 32 & 0.434578 & $[-0.0006,1.0000]$ & \thirdresult{0.832800} & 8 & 11  & $-0.0617$ \\
\bottomrule
\end{tabular}%
}
\end{table}

\begin{table}[H]
\centering
\tablefont
\caption{\textbf{Conversion efficiency.} All costs exclude target video generation. Time is in minutes. Total time is the mean across eight LoRA conversion runs and includes 96 minutes of shared model-pair preprocessing for each DART variant. Cached time subtracts this reusable cost from the total; it is a derived per-adapter cost, not a separate timing measurement. \cmark{} and \xmark{} indicate required and not required. Lower costs are better; green, yellow, and blue mark the three lowest distinct values in each cost column.}
\label{tab:app_efficiency}
\resizebox{\textwidth}{!}{%
\begin{tabular}{@{}l l ccc ccc@{}}
\toprule
\multirow{2}{*}{\textbf{Category}} & \multirow{2}{*}{\textbf{Method}} & \multicolumn{3}{c}{\textbf{Requirements}} & \multicolumn{3}{c}{\textbf{Conversion cost}} \\
\cmidrule(lr){3-5}\cmidrule(lr){6-8}
& & \textbf{Source data} & \textbf{Backward} & \textbf{Training} & \shortstack{\textbf{Forwards}\\(per LoRA)$\downarrow$} & \shortstack{\textbf{Total time}\\(min)$\downarrow$} & \shortstack{\textbf{Cached time}\\(min)$\downarrow$} \\
\midrule
Immediate & Direct & \xmark & \xmark & \xmark & \firstresult{0} & \firstresult{$<0.1$} & \firstresult{$<0.1$} \\
\midrule
\multirow{3}{*}{Training-free} & DART-W & \xmark & \xmark & \xmark & \firstresult{0} & \thirdresult{165.0} & \thirdresult{69.0} \\
& DART-C & \xmark & \xmark & \xmark & \secondresult{528} & \secondresult{158.0} & \secondresult{62.0} \\
& \textbf{DART-F} & \xmark & \xmark & \xmark & \secondresult{528} & 226.5 & 130.5 \\
\bottomrule
\end{tabular}%
}
\par\smallskip
\begin{minipage}{\textwidth}
\footnotesize
\textit{Reuse and aggregation.} The shared 96-minute setup comprises bridge construction, model-weight SVD, clustering, and coordinate-map construction for the fixed source and four-step distilled models. All three measured DART variants include this setup; additional LoRAs reuse its results. For $N$ LoRAs under the same model-pair configuration, the estimated DART-F total is $96+130.5N$ minutes, or $130.5+96/N$ minutes per LoRA.
\end{minipage}
\end{table}

\FloatBarrier

\FloatBarrier
\clearpage
\section{Additional Evaluation Results}
\label{app:additional}

\subsection{Quality dimensions and components}

The following tables provide complete quality dimensions and component ablation. Functional comparisons use the primary per-adapter FGR definition throughout.

Primary functional results use the adapter-level ratios in Table~\ref{tab:app_fgr_detail} and their macro average in Table~\ref{tab:main}. Negative values denote decreases relative to the target no-LoRA anchor.

The qualitative ablation is now reported in Figure~\ref{fig:ablation_case} in the main text, so that the visual evidence is not repeated in the appendix.
\begin{table}[!ht]
\centering
\tablefont
\caption{\textbf{Complete quality dimensions.} Subj., Bkg., Flicker, Smooth, Aesthetic, and Imaging are normalized VBench dimensions; I2V-S and I2V-B are VBench-I2V dimensions. Imaging is divided by 100 to share the zero-to-one scale. The best, second-best, and third-best entries in each numeric column are highlighted with \colorbox{RankFirst}{green}, \colorbox{RankSecond}{yellow}, and \colorbox{RankThird}{blue} backgrounds, respectively; ties share a rank.}
\label{tab:dimensions}
\resizebox{\textwidth}{!}{%
\begin{tabular*}{\textwidth}{@{\extracolsep{\fill}}l cccccccc@{}}
\toprule
\textbf{Method} & \textbf{Subj.}$\uparrow$ & \textbf{Bkg.}$\uparrow$ & \textbf{Flicker}$\uparrow$ & \textbf{Smooth}$\uparrow$ & \textbf{Aesthetic}$\uparrow$ & \textbf{Imaging}$\uparrow$ & \textbf{I2V-S}$\uparrow$ & \textbf{I2V-B}$\uparrow$ \\
\midrule
Direct & \thirdresult{0.8766} & \thirdresult{0.9218} & 0.9779 & 0.9872 & \thirdresult{0.6713} & 0.6666 & \thirdresult{0.9532} & \thirdresult{0.9580} \\
CASA & 0.8634 & 0.9153 & 0.9839 & 0.9909 & 0.6315 & \thirdresult{0.6955} & 0.9328 & 0.9409 \\
DART-W & 0.8640 & 0.9144 & \thirdresult{0.9844} & \secondresult{0.9916} & 0.6375 & 0.6871 & 0.9396 & 0.9476 \\
DART-C & \secondresult{0.9148} & \secondresult{0.9393} & \firstresult{0.9862} & \secondresult{0.9916} & \secondresult{0.6885} & \firstresult{0.6978} & \secondresult{0.9692} & \secondresult{0.9713} \\
\textbf{DART-F} & \firstresult{0.9247} & \firstresult{0.9469} & \firstresult{0.9862} & \firstresult{0.9917} & \firstresult{0.6932} & \secondresult{0.6964} & \firstresult{0.9718} & \firstresult{0.9724} \\
\bottomrule
\end{tabular*}%
}
\end{table}

\begin{table}[!ht]
\centering
\tablefont
\caption{\textbf{DART component ablation.} Transport denotes clustered coordinate transport and calibration denotes target-schedule response fitting. $Q_{\rm joint}$ is the mean of VBench-Q and I2V-Avg. Boolean entries use \cmark{} and \xmark{}. The best, second-best, and third-best numeric entries in each column are highlighted with \colorbox{RankFirst}{green}, \colorbox{RankSecond}{yellow}, and \colorbox{RankThird}{blue} backgrounds, respectively.}
\label{tab:ablation}
\resizebox{\textwidth}{!}{%
\begin{tabular}{@{}l l cc cccc@{}}
\toprule
\multirow{2}{*}{\textbf{Category}} & \multirow{2}{*}{\textbf{Variant}} &
\multicolumn{2}{c}{\textbf{Components}} & \multicolumn{4}{c}{\textbf{Quality}} \\
\cmidrule(lr){3-4}\cmidrule(lr){5-8}
& & \textbf{Transport} & \textbf{Calibration} & \textbf{$Q_{\rm joint}$}$\uparrow$ & \textbf{I2V-S}$\uparrow$ & \textbf{I2V-B}$\uparrow$ & \textbf{I2V-Avg}$\uparrow$ \\
\midrule
Baseline & Direct & \xmark & \xmark & \thirdresult{0.9029} & \thirdresult{0.9532} & \thirdresult{0.9580} & \thirdresult{0.9556} \\
\midrule
\multirow{2}{*}{Single component} & DART-W & \cmark & \xmark & 0.8950 & 0.9396 & 0.9476 & 0.9436 \\
& DART-C & \xmark & \cmark & \secondresult{0.9200} & \secondresult{0.9692} & \secondresult{0.9713} & \secondresult{0.9702} \\
\midrule
Full method & \textbf{DART-F} & \cmark & \cmark & \firstresult{0.9227} & \firstresult{0.9718} & \firstresult{0.9724} & \firstresult{0.9721} \\
\bottomrule
\end{tabular}%
}
\end{table}

\FloatBarrier

\FloatBarrier
\clearpage
\section{Limitations and Future Work}
\label{app:limitations}

Our current evaluation covers six retained adapters and three distilled targets, providing evidence across multiple customization types and target architectures but not exhaustive coverage of the broader LoRA ecosystem. The four-anchor protocol relies on task-specific functional proxies, which capture selected adapter behaviors rather than complete semantic equivalence. Response calibration uses a small probe budget to keep conversion practical; future work will investigate more diverse probe construction, adaptive probe selection, and broader prompt--reference-image coverage.
DART is target-specific, although its model-pair preprocessing is reusable across adapters and can be amortized as the adapter set grows. Future work will further extend DART to larger adapter collections, additional distillation schedules and video-model families, while developing more general functional metrics and lower-cost calibration strategies.

\end{document}